%% file: main.tex
\PassOptionsToPackage{table,dvipsnames}{xcolor} 

\documentclass{article}

\usepackage[sort&compress,numbers]{natbib}

\usepackage[final]{neurips_2025}

\usepackage{float}
\usepackage{multirow}
\usepackage{pifont}
\usepackage{listings}

\usepackage{enumitem}
\usepackage{amsmath}
\usepackage{wasysym}
\usepackage{xspace}
\usepackage[utf8]{inputenc} 
\usepackage[T1]{fontenc}    
\usepackage{hyperref}       
\usepackage{url}            
\usepackage{booktabs}       
\usepackage{amsfonts}       
\usepackage{nicefrac}       
\usepackage{microtype}      
\usepackage{tcolorbox}

\usepackage{algorithm}
\usepackage{algorithmic}

\def\eg{\emph{e.g.,}\xspace} 

\def\ie{\emph{i.e.,}\xspace} 

\def\method{\textbf{SATURN}\xspace}
\def\dataset{\textbf{SATURN-2.6k}\xspace}

\title{SATURN: \underline{SAT}-based Reinforcement Learning to \underline{U}nleash Language Model \underline{R}easo\underline{n}ing}

\author{%
  Huanyu Liu \\
  Peking University \\
  \texttt{huanyuliu@stu.pku.edu.cn} \\
  \And
  Ge Li\thanks{Corresponding author.} \\
  Peking University \\
  \texttt{lige@pku.edu.cn} \\
  \And
  Jia Li \\
  Tsinghua University \\
  \texttt{jia\_li@mail} \\
  \texttt{.tsinghua.edu.cn} \\
  \And
  Hao Zhu \\
  Peking University \\
  \texttt{zhuhao@stu.pku.edu.cn} \\
  \And
  Kechi Zhang \\
  Peking University \\
  \texttt{zhangkechi@pku.edu.cn} \\
  \And
  Yihong Dong \\
  Peking University \\
  \texttt{dongyh@stu.pku.edu.cn} \\
}

\begin{document}

\maketitle

\input{section/0Abstract}

\input{section/1Introduction}

\input{section/2SATURN}

\input{section/3Experiments}

\input{section/4Discussion}

\input{section/5Related_Work}

\input{section/6Conclusion}

\input{section/Acknowledgements}

\bibliographystyle{ACM-Reference-Format}
\bibliography{ref}

\input{section/checklist}

\input{section/appendix}

\end{document}

%% file: section/0Abstract.tex
\begin{abstract}
How to design reinforcement learning (RL) tasks that effectively unleash the reasoning capability of large language models (LLMs) remains an open question. Existing RL tasks (\eg math, programming, and constructing reasoning tasks) face three key limitations: \ding{182}~\textbf{Scalability}. They rely heavily on human annotation or expensive LLM synthesis to generate sufficient training data. \ding{183}~\textbf{Verifiability}. LLMs' outputs are hard to verify automatically and reliably. \ding{184}~\textbf{Controllable Difficulty}. Most tasks lack fine-grained difficulty control, making it challenging to train LLMs from easy to hard and progressively develop reasoning capability.

To address these limitations, we propose \method, a SAT-based RL framework that uses Boolean Satisfiability (SAT) problems to train and evaluate LLM reasoning. \method enables scalable task construction, rule-based verification, and precise difficulty control. \method designs a curriculum learning pipeline that continuously improves LLMs' reasoning capability by constructing SAT tasks of increasing difficulty and training LLMs from easy to hard. To ensure stable training, we design a principled mechanism to control difficulty transitions.

We introduce \dataset, a dataset of 2,660 SAT problems with varying difficulty. It supports the evaluation of how LLM reasoning changes with problem difficulty. We apply \method to DeepSeek-R1-Distill-Qwen and obtain \method-1.5B and \method-7B. We achieve several notable results:
\ding{182}~On SAT problems, \method-1.5B and \method-7B achieve average \texttt{pass@3} improvements of +14.0 and +28.1, respectively.
\ding{183}~On math and programming tasks, \method-1.5B and \method-7B improve average scores by +4.9 and +1.8 on benchmarks (\eg AIME, LiveCodeBench).
\ding{184}~Compared to the state-of-the-art (SOTA) approach in constructing RL tasks, \method achieves further improvements of +8.8\%.
We release the source code, data, and models to support future research at \url{https://github.com/gtxygyzb/Saturn-code}.
\end{abstract}


%% file: section/1Introduction.tex
\section{Introduction}
\label{sec:introduction}
Recently, reinforcement learning (RL) has become a promising paradigm for unleashing the reasoning capability of large language models (LLMs), particularly in math, programming, and logical reasoning (\eg OpenAI-o1~\cite{openai_o1}, DeepSeek-R1~\cite{deepseek_r1}, Kimi-k1.5~\cite{kimi_k1.5}). During the RL training process, the design of RL tasks plays a critical role~\cite{openai_o3, openai_o3_web, s1}. A well-designed RL task should elicit LLMs’ reasoning capability, fostering behaviors such as hesitation, reflection, backtracking, summarization, and verification~\cite{cot, self_verification, stepback, s1, Verify_Step}.

However, how to design RL tasks that can continuously enhance LLMs' reasoning capability remains an open question. We think a well-designed RL task for reasoning should satisfy the following three criteria:  
\ding{182} \textbf{Scalability.} RL training requires large-scale data. RL tasks should support scalable data without human annotation or expensive LLMs' synthesis.  
\ding{183} \textbf{Verifiability.} RL rewards must be unambiguously correct. The outputs of LLMs for the task should be easy to verify. 
\ding{184} \textbf{Controllable Difficulty.} Reasoning capability emerges progressively~\cite{Curriculum_Learning}. RL tasks should support the difficulty control to enable curriculum learning, allowing LLMs to gradually develop more complex reasoning skills~\cite{difficulty_controllable}.

\input{table/tasks_comparison}

Table~\ref{tab:tasks_comparison} shows the features of current mainstream RL tasks. None of them satisfy all three criteria. Existing RL tasks can be divided into two categories: (1) One category of RL tasks requires LLMs to solve math or programming problems, with rewards based on the correctness of the final answer or code~\cite{humaneval, aplhacode, mbpp, gsm8k}. However, these tasks rely on human annotation for ground-truth solutions or test cases, suffer from a lack of high-quality problems, and offer only coarse control over reasoning difficulty~\cite{aixcoder,aixcoderv2}. (2) Another category focuses on manually designed reasoning tasks~\cite{question_synthesis,wolf1,wolf2,kk2}. For instance, Logic-RL~\cite{Logic-RL} leverages natural language logic K\&K puzzles to improve LLMs' reasoning capability through RL. However, they also present limitations,  such as hard to scale up due to reliance on sampling from LLMs~\cite{Self-playing-Adversarial, question_synthesis}, hard to verify despite relying on LLMs for cross-validation~\cite{codedpo, question_synthesis, puzzbench}, and hard to control difficulty~\cite{wolf1, wolf2}.

In this paper, we aim to answer the following research question:

\begin{tcolorbox}[colback=gray!10!white, colframe=black, title=Key Question]
Can we design an RL task that supports scalability, verifiability, controllable difficulty, and enhances the reasoning capability of LLMs?
\end{tcolorbox}

\begin{figure}[h]
\centering
\includegraphics[width=1\linewidth]{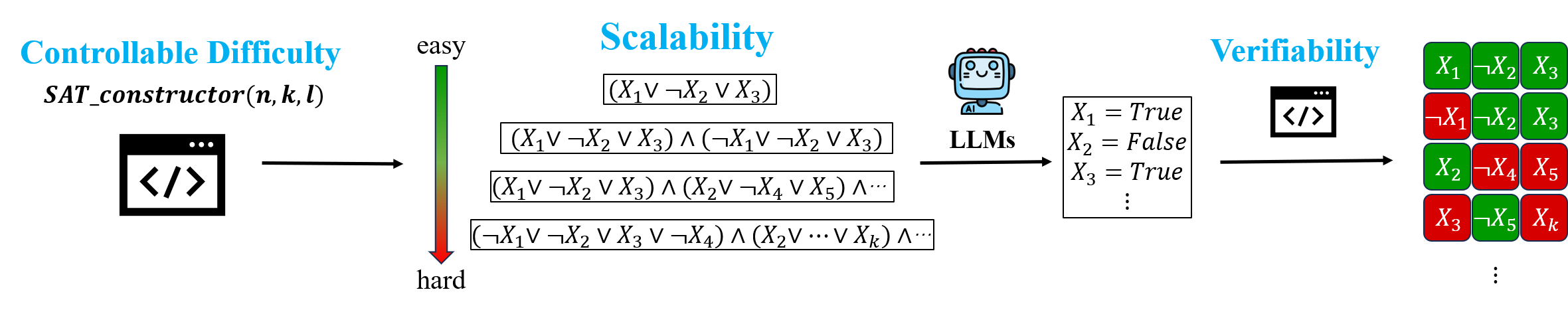}
\caption{An illustration of SAT problems and its corresponding features.}
\label{fig:sat-instance}
\end{figure}

To this end, we propose \textbf{Boolean Satisfiability (SAT) problem} as the task for RL. Figure~\ref{fig:sat-instance} shows an illustration of SAT problems and corresponding features. SAT satisfies all three desiderata we outlined earlier: \ding{182}Scalability. SAT instances can be generated programmatically at scale without human annotation or LLM synthesis, allowing for virtually unlimited training data. \ding{183} Verifiability. SAT is a well-established NP-complete problem in theoretical computer science \cite{SAT}. The correctness of a solution can be easily verified in linear time. But solving SAT problems requires complex reasoning. \ding{184} Controllable Difficulty. The difficulty of SAT instances can be precisely adjusted (\eg number of variables, clauses), making it suitable for curriculum learning. What's more, SAT serves as a universal substrate for limited forms of logical reasoning, as many problems in propositional logic, finite-domain first-order logic, and modal logic can be systematically reduced to SAT~\cite{logic1,logic2,logic3}.

Building on these advantages, we propose \emph{\underline{\textbf{SAT}}-based reinforcement learning to \underline{\textbf{U}}nleash LLMs \underline{\textbf{R}}easo\underline{\textbf{N}}ing}, or \method. \method is a multi-stage curriculum learning-based RL framework that continuously improves the reasoning capability of LLMs. \method efficiently constructs SAT tasks with controllable difficulty and organizes them into progressive stages from easy to hard, allowing LLMs to develop reasoning skills step by step. To ensure training stability and effective progression, we design a principled mechanism to control difficulty transitions based on LLMs' performance. \method enables smooth curriculum advancement and steady enhancement of reasoning capability.

We introduce the \dataset dataset, consisting of 1,500 training instances, 160 test instances at the same difficulty as the training set, and 1,000 test instances from 10 \textbf{harder} unseen difficulty levels. The test set serves as a benchmark for systematically evaluating how LLMs' reasoning capability varies with increasing SAT task difficulty. We release SAT construction scripts alongside the dataset, which enable the creation of virtually unlimited SAT instances.

We apply \method to DeepSeek-R1-Distill-Qwen-1.5B and 7B \cite{deepseek_r1}, obtaining \method-1.5B and \method-7B. Experiments show that \method effectively enhances LLMs' reasoning capability in generalizable scenarios:

\begin{itemize}[leftmargin=*]
    \item \method-1.5B and \method-7B achieve strong performance on \dataset benchmarks. On unseen harder test set, two models achieve \texttt{pass@3} improvements of +14.0 and +28.1 respectively.

    \item The reasoning capability learned from \method transfers well to math and programming tasks, bringing average improvements of +4.9 and +1.8 on popular benchmarks such as AIME \cite{aime}, AMC \cite{amc}, MATH-500 \cite{math500}, GPQA Diamond \cite{GPQA}, and LiveCodebench \cite{lcb} for two \method models.
    
    \item Compared to the prior SOTA approach (\eg Logic-RL), \method achieve average improvements of +8.8\% on math and programming tasks.

\end{itemize}

%% file: table/tasks_comparison.tex
\begin{table}[htbp]
    \centering
    \caption{The comparison between existing RL tasks and \method.}
    \label{tab:tasks_comparison}
    \resizebox{0.75 \textwidth}{!}{
    \begin{tabular}{lccc}
        \toprule
        Tasks & Scalability & Verifiability & Controllable Difficulty \\
        \midrule
        ScaleQuest \cite{question_synthesis} & \ding{55} & \ding{55} & \ding{55} \\
        GSM8K (Math) \cite{gsm8k} & \ding{55} & \ding{51} & \ding{55} \\
        LiveCodeBench \cite{lcb} & \ding{55} & \ding{51} & \ding{55} \\
        Game Werewolf \cite{wolf1, wolf2} & \ding{55} & \ding{55} & \ding{55} \\
        LMRL Gym \cite{lmrl} & \ding{55} & \ding{51} & \ding{51} \\
        SPAG \cite{Self-playing-Adversarial} & \ding{55} & \ding{51} & \ding{55} \\
        Knights\&Knaves \cite{kk2} & \ding{51} & \ding{51} & \ding{55} \\
        \midrule
        \method (Ours) & \ding{51} & \ding{51} & \ding{51} \\
        \bottomrule
    \end{tabular}}
\end{table}

%% file: section/2SATURN.tex
\section{SATURN}
\label{sec:SATURN}

\begin{figure}[htbp]
\centering
\includegraphics[width=1\linewidth]{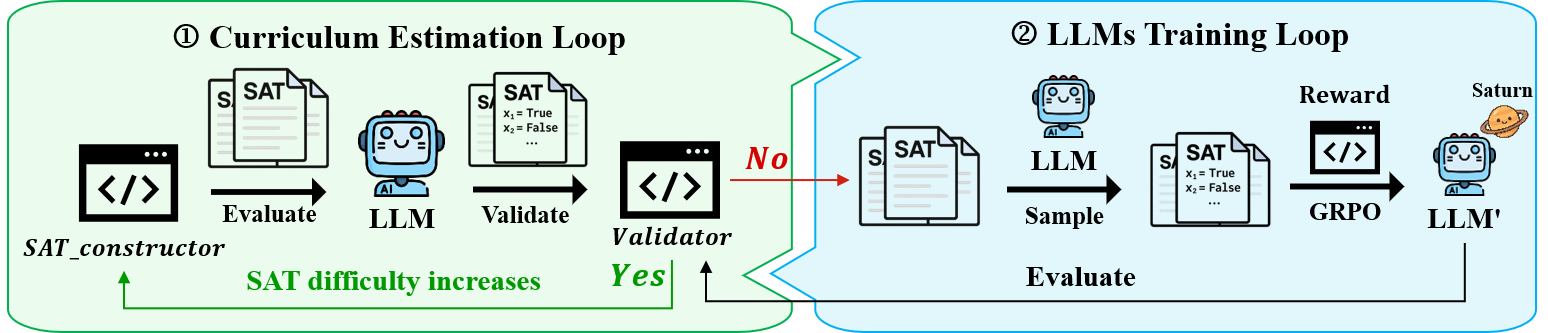}
\caption{The overall framework of \method. It alternates between two interconnected loops: 
(1) \textbf{Curriculum Estimation Loop.}
(2) \textbf{LLMs Training Loop.}
The two loops iterate until the maximum number of curriculum stages is reached.
}
\label{fig:framework}
\end{figure}

\subsection{\method Learning Loop Framework}
\label{sec:2_1}

We introduce \method, a multi-stage RL framework that leverages SAT tasks to unleash LLMs' reasoning via curriculum learning. As illustrated in Figure~\ref{fig:framework}, \method alternates between two interconnected loops: \textit{Curriculum Estimation Loop} dynamically constructs SAT instances of adjustable difficulty and evaluates LLMs’ performance to determine whether to advance the curriculum stage; \textit{LLMs Training Loop} employs RL to optimize LLMs on current difficulty SAT tasks. The curriculum loop presented in Algorithm~\ref{alg:curriculum} proceeds as follows:

\textbf{Step 1: Curriculum Estimation Loop.} Given initial SAT difficulty, \texttt{SAT\_Construction} generates a validation set of SAT instances. The LLM is evaluated on this set using the \texttt{pass@1} metric. If the performance exceeds a predefined threshold $\epsilon$, the curriculum controller advances to a harder configuration with an increased estimated difficulty. Otherwise, \method process enters \textbf{Step 2} LLMs training loop at the current SAT difficulty. This adaptive loop ensures that the LLM is always trained at the frontier of its reasoning capability, neither too easy nor too hard.

\textbf{Step 2: LLMs Training Loop.} For the current difficulty, \texttt{SAT\_Construction} generates a set of training instances that are different from the validation set. These samples are then used to train LLMs with GRPO. The reward function encourages outputs that are both logically correct and properly formatted. 
The training loop proceeds until $\texttt{pass@1} > \epsilon$ on the validation set.
After that, the process backs to \textbf{Step~1} to reassess and potentially advance the curriculum.

The two loops iterate jointly. \method process terminates when a predefined total number of iterations is reached. Importantly, \method is not designed to replace math or programming tasks, but to serve as a complementary strategy for enhancing LLMs' reasoning. In practice, \method can be integrated with math and programming tasks to enable a stronger training framework. 

\method learning loop raises three core challenges:  \ding{182} Section~\ref{sec:2_2} introduces how to construct scalable and controllable SAT instances. \ding{183} Section~\ref{sec:2_3} presents how to estimate instance difficulty for curriculum learning. \ding{184} Section~\ref{sec:2_4} explains how to train LLMs on SAT tasks with RL.



\subsection{SAT Instances Construction}
\label{sec:2_2}
In this subsection, we formalize the construction of SAT instances. A SAT problem determines whether a propositional formula can be satisfied by a Boolean truth assignment. Formally, we define a $(n, k, l)$-SAT instance in conjunctive normal form (CNF) as:  

\begin{equation}
\left\{
\begin{aligned}
\Phi =\ 
&\left(x_{a_{1,1}} \lor \neg x_{a_{1,2}} \lor \dots \lor x_{a_{1,n}} \right)  
\land \cdots 
\land \left(x_{a_{l,1}} \lor \dots \lor \neg x_{a_{l,n}} \right) \\
\text{where } &a_{i,j} \in \{1, \dots, k\}, \quad i \in [1, l]_{\mathbb{Z}}, j \in [1, n]_{\mathbb{Z}}
\end{aligned}
\right.
\end{equation}

where each clause contains exactly $n$ variables (literals), each being either $x_i$ or its negation $\neg x_i$, $k$ is the total number of variables, and $l$ is the total number of clauses. Based on the definition, we design a SAT instance constructor, \texttt{SAT\_Construction}$(n, k, l, m)$, which uniformly samples $m$ SAT instances from the space of $(n, k, l)$-SAT.  By adjusting the parameters $(n, k, l, m)$, \texttt{SAT\_Construction} enables the scalable and controllable construction of SAT instances. The design details of the constructor algorithm are provided in Appendix~\ref{app:sat_constructor}. All generated SAT instances are guaranteed to be satisfiable.



\subsection{Estimation of Task Difficulty}
\label{sec:2_3}

In this subsection, we present the estimation of SAT task difficulty for LLMs. This estimation also serves as the foundation for curriculum learning in LLMs.

As a canonical NP-complete problem \cite{SAT}, SAT admits a polynomial-time reduction from any other NP problem \cite{reduction}. SAT exhibits a known phase transition phenomenon: when the clause-to-variable ratio $\alpha_c=l/k$ approaches a critical threshold (typically near $4.26$ for 3-SAT), the probability of satisfiability drops sharply, and problem difficulty peaks. This phenomenon probably stems from \emph{replica symmetry breaking} (RSB) \cite{SAT-RSB}: near $\alpha_c$, the solution space fractures into disconnected clusters separated by energy barriers. Beyond $\alpha_c$, the space collapses, reducing complexity.

However, RSB theory is designed for heuristic SAT solvers. For humans or LLMs solving SAT problems through logical steps such as trial, verification, and reasoning, such solver-like phase transitions are hardly observable in human-like thinking processes. While any $n$-SAT ($n>3$) can be reduced to 3-SAT \cite{reduction}, they differ significantly for LLMs in terms of solution space size and token length.

Prior work \cite{sat_iclr_1} on SAT tasks for LLMs typically categorized difficulty based on phase transition points. To systematically estimate task difficulty, we define an analytical estimator of the expected solution space size. Given a $(n, k, l)$ SAT instance, its difficulty for LLMs can be approximately estimated by:

\begin{equation}
\footnotesize
D(n, k, l) = \log_2(k) + 2\log_2(l) - n + \frac{k}{n}
\label{eq:D}
\end{equation}

\begin{figure}[htbp]
\centering
\includegraphics[width=0.65\linewidth]{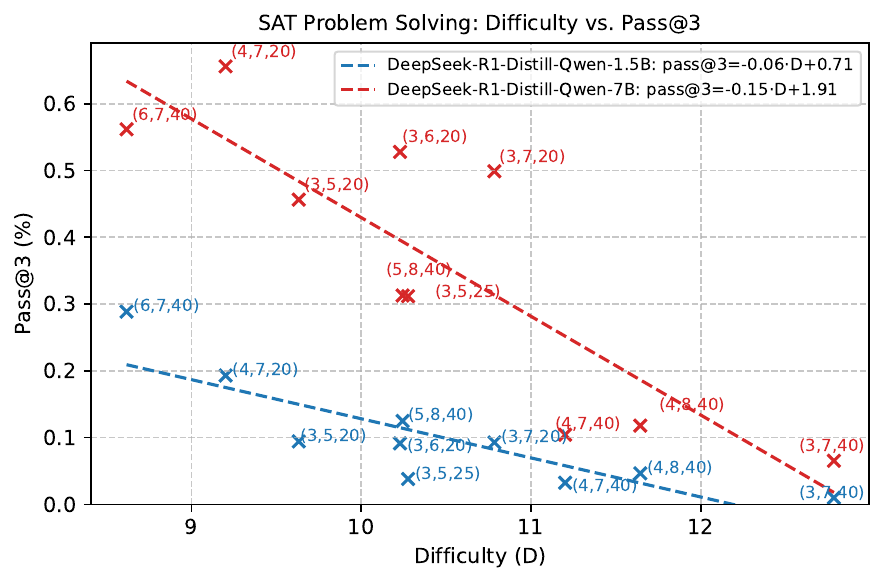}
\caption{Scatter plots of \texttt{pass@3} versus estimated difficulty $D(n, k, l)$ for different LLMs, with linear regression fits. The linear regression for two models achieve $ R^2 $ values of 0.707 and 0.724 respectively, suggesting a reasonably strong linear relationship between difficulty and \texttt{pass@3}.}
\label{fig:difficulty_vs_accuracy}
\end{figure}

Eq.~(\ref{eq:D}) provides a more controllable, fine-grained estimation of SAT task difficulty. The detailed derivation is provided in Appendix~\ref{app:sat_estimation}. To further validate Eq.~(\ref{eq:D}), we evaluate LLMs' performance on SAT instances with varying difficulty levels. As shown in Figure~\ref{fig:difficulty_vs_accuracy}, each point represents a LLM’s average \texttt{pass@3} on the same estimated difficulty instances. \texttt{Pass@3} generally decreases as $D(n, k, l)$ increases, suggesting that our estimation aligns with the solvability trends observed in practical LLMs. Stronger LLMs maintain higher \texttt{pass@3}, while weaker LLMs exhibit lower scores overall. The validity of the estimation in Eq.~(\ref{eq:D}) is further confirmed by ablation experiments, as detailed in Appendix~\ref{app:sat_estimation}.

\subsection{Reinforcement Learning with GRPO}
\label{sec:2_4}

In this subsection, we introduce the single-stage RL training for given $(n, k, l)$-difficulty tasks. RL can further improve LLMs' generalization by directly optimizing policy gradients over diverse reasoning trajectories~\cite{deepseek_r1}. Given the SAT tasks, we then train LLMs using the original sample-level GRPO to optimize the policy $\pi_\theta$ with KL divergence penalty. The GRPO objective function is defined as: 

\begin{equation}
\footnotesize
\begin{aligned}
\mathcal{L}_{\text{GRPO}}(\theta) =& 
 \mathbb{E}{[q \sim P(Q), \{o_i\}_{i=1}^G \sim \pi_{\theta_{old}}(O|q)]}  \\
 &\frac{1}{G}\sum_{i=1}^G\frac{1}{|o_i|} \sum_{t=1}^{|o_i|} \left\{ 
\min \left[ r_{i,t}(\theta) \hat{A}_{i,t}, 
\text{clip} \left( r_{i,t}(\theta), 1 - \epsilon, 1 + \epsilon \right)  \hat{A}_{i,t} \right] 
- \beta \mathbb{D}_{\text{KL}}\left[\pi_{\theta} \parallel \pi_{\text{ref}}\right]
\right\} \\
\text{where }
r_{i,t}(\theta) &= \frac{\pi_{\theta}(o_{i,t} \mid q, o_{i,<t})}{\pi_{\theta_{\text{old}}}(o_{i,t} \mid q, o_{i,<t})}, \quad \hat{A}_{i,t} = \frac{r_i - \text{mean}(\mathbf{r})}{\text{std}(\mathbf{r})}
\end{aligned}
\end{equation}

where $q$ denotes a SAT instance, $o_i$ is the reasoning trajectory generated by the policy $\pi_\theta$, and $G$ groups SAT instances with identical $(n,k,l)$ parameters. A simple yet effective reward scheme~\cite{deepseek_r1} is designed that combines a \textit{format reward} and a \textit{correctness} reward. Specifically, $r_i = -1$ if an output is invalid (\ie missing the \texttt{\textbackslash boxed\{\}} wrapper); $r_i = 0$ for well-formatted but incorrect answers; and $r_i = 1$ only when both the format and the answer are correct. Here, an answer is considered correct if it passes a verifier and represents a full satisfying assignment. Training schedule and hyperparameter settings are detailed in Appendix~\ref{app:sat_training}. And the SAT prompt template is shown in Appendix~\ref{app:eval_prompts}.

%% file: section/3Experiments.tex
\section{Experiments}
\label{sec:experiments}

We apply \method to DeepSeek-R1-Distill-Qwen-1.5B and 7B, obtaining \method-1.5B and \method-7B. To evaluate the effectiveness of \method, we conduct a large-scale study to evaluate both models. In this section, we introduce our research questions (RQs), benchmarks, baselines, and evaluation metrics. For each RQ, we present the corresponding experimental design, results, and analysis in separate subsections.

\subsection{Research Questions}
\label{sec:experiment:RQs}

Our study aims to answer the following research questions:

\textbf{RQ1: How much improvement does \method achieve in solving SAT tasks?} We evaluate \method-1.5B and \method-7B performance on \dataset with different difficulty levels.

\textbf{RQ2: How effectively does \method generalize to math and programming tasks?} To evaluate the transferability of reasoning capabilities learned by \method, we evaluate the performance of LLMs on math and programming benchmarks and compare them with current SOTA LLMs.

\textbf{RQ3: How does \method compare to prior RL tasks?} To explore the relationship between \method and existing RL tasks, RQ3 investigates whether \method can (1) serve as a complementary task to math and programming, and (2) outperform other constructing RL tasks.

\textbf{RQ4: How does \method affect LLMs reasoning trajectory?} RQ4 explores whether \method influences the reasoning patterns of LLMs, particularly in terms of response length and the capability of verification. We investigate whether the reasoning improvements observed in SAT tasks generalize to math and programming.

\subsection{Experimental Setup}
\label{sec:setup}

\paragraph{\method Hyperparameters}
For \method-1.5B and \method-7B, we set the initial SAT instance parameters \((n, k, l)\) to \((3, 5, 5)\) and \((3, 5, 13)\), respectively. In \emph{Curriculum Estimation Loop},  the \(\epsilon\) threshold is set to 0.5 for the 1.5B model and 0.75 for the 7B model. In \emph{LLMs Training Loop}, we evaluate the \texttt{pass@k} with a step size of 250 training samples. The total number of curriculum iterations is set to 2. Detailed hyperparameters are provided in Appendix~\ref{app:saturn_pseudo}. Ablation studies in Appendix~\ref{app:ablation} demonstrate the necessity of curriculum learning and the effectiveness of hyperparameters on SAT difficulty, thresholds, step sizes, etc.

\paragraph{Benchmarks.} 
\ding{182} Building upon \texttt{SAT\_Construction} tool and difficulty estimation, we release \dataset, a curated benchmark designed to evaluate LLMs' reasoning capability across varying complexity. \dataset consists of 1,500 training instances and 160 test instances that share the same estimated difficulty level. To assess performance under increasing task complexity, \dataset further includes 1,000 test instances from 10 unseen harder difficulty levels, with 100 instances per level. These levels are selected based on our difficulty estimation $D(n, k, l)$, enabling a systematic analysis of how LLM performance changes as problem difficulty increases. Additionally, custom datasets of desired difficulty can be constructed using our open-sourced \texttt{SAT\_Construction} tool. 
\ding{183} For math and programming tasks, following DeepSeek-AI \cite{deepseek_r1}, we use \textbf{AIME 24/25} \cite{aime}, \textbf{AMC 22/23} \cite{amc}, \textbf{MATH-500} \cite{math500}, \textbf{GPQA Diamond} \cite{GPQA}, and \textbf{LiveCodeBench} \texttt{v4\_v5} subset \cite{lcb}.


\paragraph{Baseline Model.} We conduct evaluations against several 1.5B and 7B parameter reasoning models as the baselines, which include DeepSeek-R1-Distill-Qwen-1.5B \& 7B \cite{deepseek_r1}, Still-3-1.5B-Preview \cite{still3}, s1.1-1.5B \& 7B \cite{s1}, z1-7B \cite{z1}, OpenThinker-7B \cite{openthoughts}, and DeepScaleR-1.5B-Preview 
\cite{deepscaler2025}. In addition, we include a supervised fine-tuning (SFT)-only baseline trained on the Math training dataset \cite{math500}, which provides step-by-step problem reasoning trajectories. We randomly select the most difficult Level-5 1,000 problems from training set for one epoch of SFT, following the same training template as DeepSeek-R1-Distill-Qwen.  With the same dataset size, our setup enables a fair comparison between SFT and RL on SAT tasks.

\paragraph{Evaluation Metrics.} Following DeepSeek-AI \cite{deepseek_r1}, we use \texttt{pass@k} as the evaluation metric. \texttt{Pass@k} assesses the probability that at least one correct solution is generated within \( k \) attempts. For SAT problems, we evaluate  $ \texttt{pass@k} \in \{1, 3, 5, 7, 10\} $ and sample 12 times per problem. 
For math and programming benchmarks, we use \texttt{pass@1}, following a context length of 32,768 and temperature = 0.6. More evaluation hyperparameters are provided in Appendix~\ref{app:eval_hyperparams}. All experiments are conducted on NVIDIA 8×A100 (40GB) GPUs. Specific prompts are detailed in Appendix~\ref{app:eval_prompts}.

\subsection{RQ1: \method Substantially Improves Performance on SAT Tasks}
\label{sec:rq1}

We evaluate the performance of \method-1.5B and \method-7B on SAT tasks using \dataset test set. Specifically, the evaluation involves unseen SAT instances that were not included in the training data. The results, presented in Table~\ref{tab:rq1_sat}, \ref{app:full_sat_pass1}--\ref{app:full_sat_pass10}, and detailed in Appendix~\ref{app:rq1}, demonstrate the performance of LLMs across different SAT difficulties.

\input{table/RQ1_sat}

\textbf{\method substantially improves LLM performance on SAT tasks across varying difficulty levels.} On the difficulty SAT-(3,5,5), \method-1.5B improves \texttt{pass@1} from 36.7 to 59.7 at Iteration-1, and further to 70.3 at Iteration-2, achieving a total gain of +33.6. On the unseen harder test set (Table~\ref{app:full_sat_pass3}), \method-1.5B improves average \texttt{pass@3} from 10.1 to 24.2, while \method-7B improves from 36.1 to 64.2. On average, these models achieve \texttt{pass@3} improvements of +14.0 and +28.1 respectively, confirming that \method effectively enhances LLM reasoning across both seen and unseen SAT difficulties.

\subsection{RQ2: \method Demonstrates Strong Generalization to Math and Programming}
\label{sec:rq2}

We assess whether the reasoning capability learned by \method generalizes to math and programming tasks. We evaluate \method-1.5B and \method-7B on a range of reasoning benchmarks. The results shown in Table~\ref{tab:math_program} provide a detailed comparison.

\input{table/math_program}

\textbf{\ding{182} \method shows strong generalization to math and programming tasks.} On the AIME 24/25 benchmark, \method-1.5B outperforms z1-7B by 8.3 and s1.1-7B by 21.7. Similarly, \method-7B achieves a strong improvement on the Math500 dataset, increasing from 93.2 to 95.0. On LiveCodeBench, it improves from 35.4 to 37.7. On average, \method-1.5B improves by +4.9, and \method-7B improves by +1.8 across these benchmarks. These results highlight that \method enhances the reasoning performance of LLMs across various math and programming tasks, demonstrating strong generalization of the learned reasoning capabilities from SAT.

\textbf{\ding{183} \method outperforms SFT on broader benchmarks.} Consistent with the observations in \emph{SFT Memorizes, RL Generalizes}~\cite{SFT-RL}, SFT improves performance on math-focused benchmarks (AIME, AMC, and Math500) that are similar to its supervised training domain. However, on LiveCodeBench, SFT drops from 16.4 to 14.6, exhibiting an \emph{alignment tax}~\cite{alignment_tax}, where specializing on a narrow domain compromises performance on other tasks. In contrast, \method improves performance across all benchmarks, with \method-1.5B reaching 17.4 on LiveCodeBench. Averaging across all benchmarks, \method-1.5B outperforms the SFT counterpart by 3.3, demonstrating that \method generalizes effectively.


\subsection{RQ3: \method Serves as a Complement and Further Enhances LLM Reasoning}
\label{sec:rq3}

RQ3 studies the relationship between \method and existing RL tasks. Beyond the DeepSeek-R1-Distill-Qwen-7B, we introduce two additional models: Qwen2.5-7B-Instruct-1M \cite{qwen2.5,qwen2.5-1m} following Logic-RL \cite{Logic-RL} settings, and DeepScaleR-1.5B-Preview \cite{deepscaler2025}, which is further RL trained on 40k math and programming examples from DeepSeek-R1-Distill-Qwen-7B. We compare \method against several prior constructing RL task approaches, including Logic-RL \cite{Logic-RL}, SPGA \cite{Self-playing-Adversarial}, and ScaleQuest \cite{question_synthesis}, which represent strong baselines. Each approach is applied to different models for comparison. Results are summarized in Table~\ref{tab:math_program_extended}.

\input{table/RL_task}

\textbf{\ding{182} \method serves as a strong complement to math and programming.} On DeepScaleR-1.5B-Preview—despite being further RL trained with 40k math and programming  examples—\method still brings additional improvements, raising the average score from 49.9 to 52.3. Notably, it improves AIME by +5.0 and GPQA-D by +6.0. \textbf{\ding{183} \method outperforms prior constructing RL task approaches across multiple models.} On Qwen2.5-7B-Instruct-1M, \method uses only 1k training examples but improves the average score from 32.5 to 36.2, achieving a relative improvement of +8.8\% over Logic-RL trained with 5k examples.  These results indicate that \method not only complements math and programming tasks, but also provides greater improvements compared to other constructing RL task approaches.


\subsection{RQ4: \method Enhances Self-verification in LLMs' Reasoning Trajectories}
\label{sec:rq4}

\begin{figure}[htbp]
\centering
\includegraphics[width=0.5\linewidth]{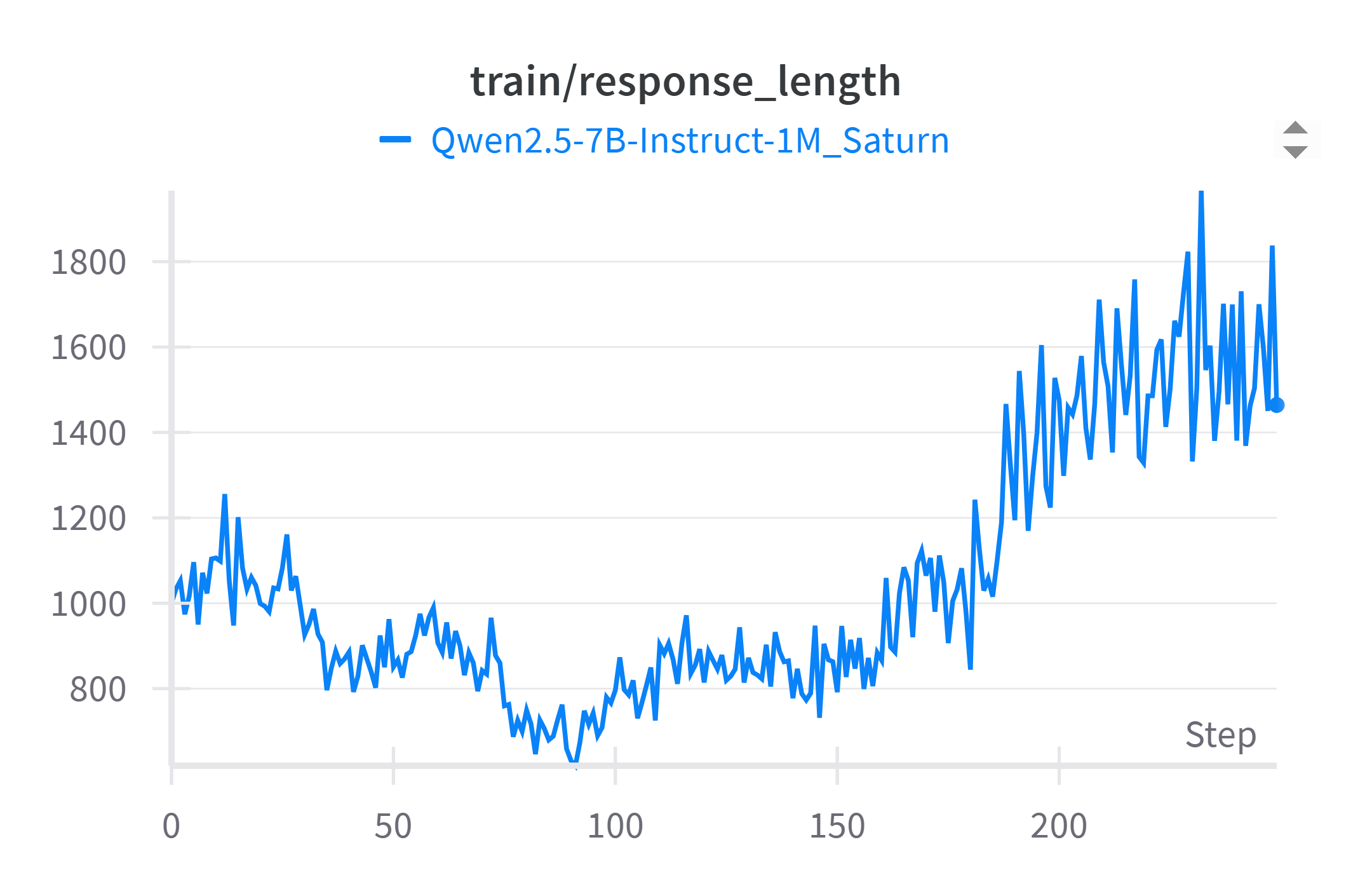}
\caption{Response length trend during \method training on Qwen2.5-7B-Instruct-1M.}
\label{fig:rq4_length}
\end{figure}

RQ4 investigates whether \method affects LLMs' reasoning behavior. On Qwen2.5-7B-Instruct-1M, we observe a gradual increase in response length during training, as illustrated in Figure~\ref{fig:rq4_length}, replicating the lengthening phenomenon reported in the R1 and Logic-RL~\cite{deepseek_r1, Logic-RL}.

\begin{figure}[t]
\centering
\includegraphics[width=\linewidth]{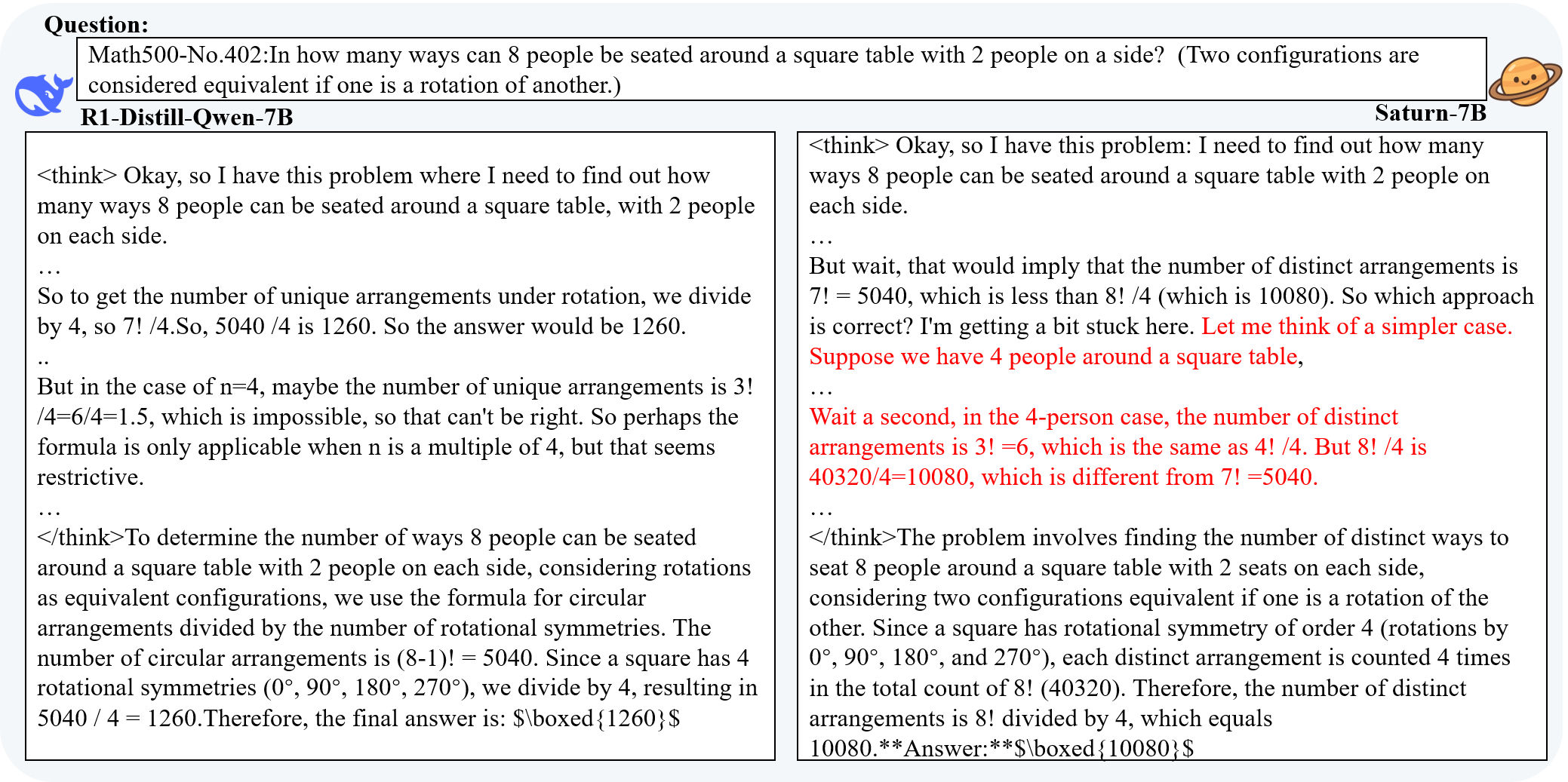}
\caption{Comparison of reasoning trajectories on Math 500 Question No.402 between two models.}
\label{fig:rq4_case_study_402}
\vspace{-0.3cm}
\end{figure}

To examine whether such reasoning patterns generalize, we present case studies across SAT and math domains. Figure~\ref{fig:rq4_case_study_sat} shows that solving SAT variables requires rechecking all clauses, naturally encouraging self-verification. In Figure~\ref{fig:rq4_case_study_402}, \method-7B verifies intermediate conclusions within a small scenario and successfully chooses the correct solution path. In contrast, the baseline model reaches a wrong answer and skips verification, even when inconsistencies are detected. 

Recent studies~\cite{Cognitive, Meta-Abilities} identify core behaviors shared by expert human reasoners and LLMs, such as \textbf{verification} and \textbf{backtracking}. These behaviors are domain-agnostic and provide fundamental reasoning patterns applicable to a wide range of tasks. In line with these findings, \method reinforces similar behaviors during SAT solving, leading to more structured reasoning trajectories. More reasoning trajectories are provided in Appendix~\ref{app:case_study} to illustrate how SATURN works. These results suggest that the self-verification patterns learned from SAT transfer well to math and programming tasks, improving reasoning robustness and reliability.

%% file: table/RQ1_sat.tex
\begin{table}[htbp]
    \centering
    \caption{
        Performance (\textbf{pass@k}, in \%) on \dataset test set across different difficulty levels.
    }
    \label{tab:rq1_sat}
    \resizebox{\textwidth}{!}{
    \begin{tabular}{lccccc|ccccc}
        \toprule
        \multirow{2}{*}{\textbf{Model}} 
        & \multicolumn{5}{c|}{\textbf{SAT-(3,5,5)}} 
        & \multicolumn{5}{c}{\textbf{SAT-(3,5,8)}} \\
        \cmidrule(lr){2-6} \cmidrule(lr){7-11}
        & @1 & @3 & @5 & @7 & @10 & @1 & @3 & @5 & @7 & @10 \\
        \midrule
        DeepSeek-R1-Distill-Qwen-1.5B & 36.7 & 71.7 & 85.4 & 91.7 & 96.2 & 20.3 & 47.6 & 63.6 & 73.4 & 81.9 \\
        \method-1.5B-Iteration-1 & 59.7 & 90.4 & 97.1 & 99.1 & 99.8 & 41.0 & 74.0 & 85.6 & 91.1 & 95.6 \\
        \rowcolor[rgb]{ .741,  .843,  .933} \method-1.5B-Iteration-2 & \textbf{70.3} & \textbf{95.9} & \textbf{99.0} & \textbf{99.7} & \textbf{99.9} & \textbf{47.0} & \textbf{82.6} & \textbf{93.9} & \textbf{98.0} & \textbf{99.8} \\
        \midrule
        \multirow{2}{*}{\textbf{Model}} 
        & \multicolumn{5}{c|}{\textbf{SAT-(3,5,13)}} 
        & \multicolumn{5}{c}{\textbf{SAT-(3,5,15)}} \\
        \cmidrule(lr){2-6} \cmidrule(lr){7-11}
        & @1 & @3 & @5 & @7 & @10 & @1 & @3 & @5 & @7 & @10 \\
        \midrule
        DeepSeek-R1-Distill-Qwen-7B & 53.9 & 86.2 & 94.2 & 97.3 & 99.3 & 39.3 & 74.9 & 88.3 & 94.3 & 98.3 \\
        \method-7B-Iteration-1 & 73.0 & 96.1 & 98.9 & 99.7 & 99.9 & 65.7 & 91.8 & 96.8 & 98.7 & 99.7 \\
        \rowcolor[rgb]{ .741,  .843,  .933} \method-7B-Iteration-2 & \textbf{89.5} & \textbf{99.0} & \textbf{99.9} & \textbf{100.0} & \textbf{100.0} & \textbf{85.4} & \textbf{98.3} & \textbf{99.8} & \textbf{99.9} & \textbf{100.0} \\
        \bottomrule
    \end{tabular}
    }
\end{table}

%% file: table/math_program.tex
\begin{table}[htbp]
    \centering
    \caption{Performance comparison on math and programming Benchmarks}
    \label{tab:math_program}
    \resizebox{1.0 \textwidth}{!}{
    \begin{tabular}{lcccccc}
    \toprule
    \textbf{Model} & \textbf{AIME 24/25} & \textbf{AMC 22/23} & \textbf{Math500} & \textbf{GPQA-D} & \textbf{LiveCodeBench} & \textbf{Avg.} \\
    \midrule
    GPT-4o (Aug'24)     & 11.7 & - & 79.5 & 52.1 & 31.7 & - \\
    Claude 3.5 Sonnet (Oct '24)     & 15.7 & - & 77.1 & 59.9 & 38.1 & - \\
    \midrule
    s1.1-1.5B           & 1.7  & 25.3 & 42.2 & 29.3 & 2.2  & 20.1 \\
    Still-3-1.5B-Preview & 23.3 & 74.7 & 84.6 & 34.8 & 17.1 & 46.9 \\
    DeepSeek-R1-Distill-Qwen-1.5B    & 21.6 & 65.1 & 83.6 & 30.3 & 16.4 & 43.4 \\
    \quad + SFT & 25.0	& 68.7 & 82.0 & 34.3 & 14.6 & 44.9 \\
    \rowcolor[rgb]{ .741,  .843,  .933} \method-1.5B  & \textbf{28.3} & \textbf{73.5} & \textbf{84.6} & \textbf{37.4} & \textbf{17.4} & \textbf{48.2} \\
    \midrule
    z1-7B               & 8.3  & 39.8 & 74.2 & 35.4 & 19.3 & 35.4 \\
    s1.1-7B               & 21.7 & 61.4 & 80.8 & 43.4 & 12.8 & 44.0 \\
    OpenThinker-7B      & 26.7 & 53.0 & 88.6 & 42.9 & 21.5 & 46.5 \\
    DeepSeek-R1-Distill-Qwen-7B      & \textbf{50.0} & 80.7 & 93.2 & 49.0 & 35.4 & 61.7 \\
    \rowcolor[rgb]{ .741,  .843,  .933} \method-7B  & 48.3 & \textbf{85.5} & \textbf{95.0} & \textbf{50.5} & \textbf{37.7} & \textbf{63.4} \\
    \bottomrule
    \end{tabular}}
\end{table}

%% file: table/RL_task.tex
\begin{table}[htbp]
    \centering
    \caption{Comparison of \method and prior approaches across various LLMs.}
    \label{tab:math_program_extended}
    \resizebox{1.0 \textwidth}{!}{
    \begin{tabular}{lcccccc}
    \toprule
    \textbf{Model} & \textbf{AIME 24/25} & \textbf{AMC 22/23} & \textbf{Math500} & \textbf{GPQA-D} & \textbf{LiveCodeBench} & \textbf{Avg.} \\
    \midrule
    SPGA-3 (82k)                & 0.0 & 3.6  & 7.2  & 24.7 & 0.0  & 7.1 \\
    ScaleQuest (25k)            & 6.7 & 45.8 & 74.6 & 31.3 & 7.9  & 33.3 \\
    Qwen2.5-7B-Instruct-1M & 5.0 & 41.0 & 74.4 & 32.3 & 9.8 & 32.5 \\
    \quad + Logic-RL (5k)      & 6.7 & \textbf{49.4} & 72.0 & 29.3 & 9.0 & 33.3 \\
    \rowcolor[rgb]{ .741,  .843,  .933} \textbf{\quad + Saturn (1k)}         & \textbf{10.0} & 47.0 & \textbf{74.8} & \textbf{37.9} & \textbf{11.3} & \textbf{36.2} \\
    
    \midrule
    DeepSeek-R1-Distill-Qwen-7B & \textbf{50.0} & 80.7 & 93.2 & 49.0 & 35.4 & 61.7 \\
    \quad + Logic-RL (5k)            & \textbf{50.0} & 80.7 & 93.4 & \textbf{52.0} & 35.7 & 62.4 \\
    \rowcolor[rgb]{ .741,  .843,  .933} \textbf{\quad + Saturn (1k)}              & 48.3 & \textbf{85.5} & \textbf{95.0} & 50.5 & \textbf{37.7} & \textbf{63.4} \\
    \midrule
    
    DeepScaleR-1.5B-Preview & 30.0 & 74.7 & 87.8 & 37.4 & 19.8 & 49.9 \\
    \quad + Logic-RL (5k) & 28.3   & \textbf{77.1}    & 86.4    & 35.9    & 20.7    & 49.7 \\
    \rowcolor[rgb]{ .741,  .843,  .933} \textbf{\quad + Saturn (0.5k)} & \textbf{35.0}    & 73.5    & \textbf{88.6}    & \textbf{43.4}    & \textbf{21.0}    & \textbf{52.3} \\
    \bottomrule
    \end{tabular}}
\end{table}

%% file: section/4Discussion.tex
\section{Discussion}
\label{sec:discussion}

\subsection{Limitations of Reasoning Capability Learned from \method}

During curriculum learning, we observed that as the number of training iterations increases, the improvements in math and programming tasks tend to plateau, which is consistent with the findings in Logic-RL. Detailed evaluation results are provided in Table~\ref{tab:multi_stage}. This plateau may stem from several factors:
\ding{182} \textbf{Knowledge limitations.} \method improves formal logical reasoning but does not provide domain-specific knowledge supervision. This limits its effectiveness in tasks requiring mathematical or algorithmic knowledge.
\ding{183} \textbf{Context window bottlenecks.} SAT problems are NP-complete tasks, and the required reasoning length grows exponentially with increasing problem difficulty. This leads to bottlenecks in the model’s capability to handle increasingly complex tasks.
\ding{184} \textbf{Limited plasticity and forgetting.} Model plasticity and catastrophic forgetting are known limitations that hinder further improvements with additional training stages \cite{forgetting1,forgetting2}.

\input{table/multi_stage}

\subsection{Potential of \method on Stronger Models}

To explore the potential of \method to stronger models, we evaluate frontier LLMs on SAT tasks using the extended SAT instances. Results are shown in Appendix~\ref{app:stronger}. Although these LLMs exhibit stronger performance, they still make common errors such as hallucinating clauses, confidently committing to incorrect decisions, or failing to apply basic logical rules. Even more advanced LLMs still struggle to solve complex SAT problems. We believe \method remains a promising approach for enhancing reasoning in stronger LLMs. With sufficient computation, \method can offer a scalable, verifiable, and controllable path to further improve reasoning capabilities.







%% file: table/multi_stage.tex
\begin{table}[htbp]
    \centering
    \caption{Performance of multi-Stage \method iterations on math and programming benchmarks}
    \label{tab:multi_stage}
    \resizebox{1.0 \textwidth}{!}{
    \begin{tabular}{lcccccc}
    \toprule
    \textbf{Model} & \textbf{AIME 24/25} & \textbf{AMC 22/23} & \textbf{Math500} & \textbf{GPQA-D} & \textbf{LiveCodeBench} & \textbf{Avg.} \\
    \midrule
    DeepSeek-R1-Distill-Qwen-1.5B    & 21.6 & 65.1 & 83.6 & 30.3 & 16.4 & 43.4 \\
    
    \method-1.5B-Iteration-1 & 26.7 & 68.6 & 85.0 & 33.3 & 16.9 & 46.1 \\
    
    \method-1.5B-Iteration-2  & 28.3 & \textbf{73.5} & 84.6 & \textbf{37.4} & \textbf{17.4} & \textbf{48.2} \\

    \method-1.5B-Iteration-3 & \textbf{28.3} & 66.3 & \textbf{85.8} & 36.9 & 16.7 & 46.9 \\
    
    \midrule
    
    DeepSeek-R1-Distill-7B      & \textbf{50.0} & 80.7 & 93.2 & 49.0 & 35.4 & 61.7 \\
    
    \method-7B-Iteration-1 & 48.3 & 83.1 & 94.6 & 50.5 & 36.6 & 62.6 \\
    
    \method-7B-Iteration-2  & 48.3 & 85.5 & \textbf{95.0} & 50.5 & 37.7 & 63.4 \\

    \method-7B-Iteration-3 & 46.7 & \textbf{87.9} & 93.2 & \textbf{58.1} & \textbf{38.1} & \textbf{64.8} \\
    
    \bottomrule
    \end{tabular}}
\end{table}

%% file: section/5Related_Work.tex
\section{Related Work}
\label{sec:related_work}

\subsection{Constructing Reasoning Tasks for Reinforcement Learning}

Several works have explored constructing reasoning tasks to improve the reasoning capability of LLMs. Logic-RL~\cite{Logic-RL} and LMRL Gym \cite{lmrl} train LLMs on natural language logic puzzles but lack scalability due to their limited puzzle set. ScaleQuest~\cite{question_synthesis}, Entity-Deducing Game~\cite{20_Question}, and K\&K~\cite{kk2} propose automatic generation of constructing questions, but rely on LLM sampling or handcrafted templates, making large-scale generation costly and hard to verify. CodeDPO~\cite{codedpo} and PuzzBench~\cite{puzzbench} employ LLM-based verification, which may fail silently and cannot ensure correctness. Wolf Game~\cite{wolf1,wolf2} focus on multi-step logic reasoning but offer no control over task difficulty, limiting their support for curriculum learning. Overall, these tasks fall short in scalability, verifiability, or controllable difficulty. See Appendix~\ref{app:related_work} for detailed comparisons.

\subsection{SAT-Based Evaluation of LLM Reasoning Capability}

Recent studies have evaluated the reasoning capability of LLMs on SAT problems. Most of these works focus on analyzing model behavior around the SAT phase transition~\cite{sat_iclr_1, sat_iclr_solver,sat_blog}, where problem hardness peaks. However, the phase transition theory is originally designed for heuristic SAT solvers and does not align well with the reflective and verification-based reasoning processes of humans or LLMs. These studies also lack a fine-grained scalable difficulty framework and typically divide difficulty based on the phase transition threshold. They are further limited to supervised fine-tuning and do not consider large reasoning models with long-CoT reasoning capability trained via RL. Our work addresses these limitations by building a progressive evaluation and curriculum learning pipeline, enabling precise difficulty control and the generalization of LLMs.



%% file: section/6Conclusion.tex
\section{Conclusion and Future Work}
\label{sec:conclusion}

We present \method, a SAT-based RL framework for unleashing and evaluating the reasoning capability of LLMs. By leveraging SAT’s \textbf{scalability}, \textbf{verifiability}, and \textbf{controllable difficulty}, \method addresses key limitations of existing RL tasks. It constructs a multi-stage curriculum to gradually enhance reasoning, and introduces the \dataset benchmark for controlled evaluation. Applied to DeepSeek-R1-Distill-Qwen, \method produces \method-1.5B and \method-7B, which show strong gains on unseen SAT tasks and generalize well to math and programming benchmarks.

In future work, we plan to: (1) apply \method to larger-scale LLMs, (2) break the existing paradigm's reliance on human-annotated data and explore new paths toward building LLMs with continuous self-evolution capabilities.

%% file: section/Acknowledgements.tex
\section*{Acknowledgements}

This research is supported by the National Key R\&D Program under Grant No. 2023YFB4503801, the National Natural Science Foundation of China under Grant No. 62192733, 62192730, 62192731, the Major Program (JD) of Hubei Province (No.2023BAA024), and the Beijing Natural Science Foundation under Grant No. 4264107.

%% file: section/checklist.tex
\section*{NeurIPS Paper Checklist}

\begin{enumerate}

\item {\bf Claims}
    \item[] Question: Do the main claims made in the abstract and introduction accurately reflect the paper's contributions and scope?
    \item[] Answer: \answerYes{} 
    \item[] Justification: The abstract and introduction clearly state the main contributions, including the curriculum learning framework \method and the \dataset dataset. All supported by experimental results.

\item {\bf Limitations}
    \item[] Question: Does the paper discuss the limitations of the work performed by the authors?
    \item[] Answer: \answerYes{} 
    \item[] Justification: We discuss the limitations of \method in Section~\ref{sec:discussion}.

\item {\bf Theory assumptions and proofs}
    \item[] Question: For each theoretical result, does the paper provide the full set of assumptions and a complete (and correct) proof?
    \item[] Answer: \answerYes{} 
    \item[] Justification: Any informal proof provided in the core of the paper is complemented by formal proofs in the appendix, including the rigorous derivation of SAT estimation.

    \item {\bf Experimental result reproducibility}
    \item[] Question: Does the paper fully disclose all the information needed to reproduce the main experimental results of the paper to the extent that it affects the main claims and/or conclusions of the paper (regardless of whether the code and data are provided or not)?
    \item[] Answer: \answerYes{} 
    \item[] Justification: We release all code, models, datasets, and experimental results to ensure full reproducibility.

\item {\bf Open access to data and code}
    \item[] Question: Does the paper provide open access to the data and code, with sufficient instructions to faithfully reproduce the main experimental results, as described in supplemental material?
    \item[] Answer: \answerYes{} 
    \item[] Justification: We release all code, models, datasets, and experimental results, along with detailed information on the execution environment to ensure full reproducibility.


\item {\bf Experimental setting/details}
    \item[] Question: Does the paper specify all the training and test details (e.g., data splits, hyperparameters, how they were chosen, type of optimizer, etc.) necessary to understand the results?
    \item[] Answer: \answerYes{} 
    \item[] Justification: We provide all important parameters needed to understand the results in the appendix, and the code scripts include complete training and test details.

\item {\bf Experiment statistical significance}
    \item[] Question: Does the paper report error bars suitably and correctly defined or other appropriate information about the statistical significance of the experiments?
    \item[] Answer: \answerNo{} 
    \item[] Justification: Following current evaluation practices in recent large model benchmarks, we report the unbiased estimator of the \texttt{pass@k} metric. This expectation-based metric is designed to reflect statistical significance across multiple samples and inherently captures sampling variability. Although we do not report explicit error bars, \texttt{pass@k} inherently reflects statistical reliability. Further details are provided in Appendix~\ref{app:eval_hyperparams}.

\item {\bf Experiments compute resources}
    \item[] Question: For each experiment, does the paper provide sufficient information on the computer resources (type of compute workers, memory, time of execution) needed to reproduce the experiments?
    \item[] Answer: \answerYes{} 
    \item[] Justification: We provide details on the GPU resources used for experiments as well as information on all data storage.
    
\item {\bf Code of ethics}
    \item[] Question: Does the research conducted in the paper conform, in every respect, with the NeurIPS Code of Ethics \url{https://neurips.cc/public/EthicsGuidelines}?
    \item[] Answer: \answerYes{} 
    \item[] Justification: We have reviewed and ensured that all research conducted in this paper fully conforms to the NeurIPS Code of Ethics.

\item {\bf Broader impacts}
    \item[] Question: Does the paper discuss both potential positive societal impacts and negative societal impacts of the work performed?
    \item[] Answer: \answerNA{} 
    \item[] Justification: This paper focuses on improving LLMs' reasoning capability through \method curriculum learning framework, without direct deployment or application, and thus does not raise societal impacts concerns.
    
\item {\bf Safeguards}
    \item[] Question: Does the paper describe safeguards that have been put in place for responsible release of data or models that have a high risk for misuse (e.g., pretrained language models, image generators, or scraped datasets)?
    \item[] Answer: \answerNA{} 
    \item[] Justification: This paper poses no foreseeable risks of misuse, as it does not involve the release of high-risk models or datasets.

\item {\bf Licenses for existing assets}
    \item[] Question: Are the creators or original owners of assets (e.g., code, data, models), used in the paper, properly credited and are the license and terms of use explicitly mentioned and properly respected?
    \item[] Answer: \answerYes{} 
    \item[] Justification: We properly credit all used assets in the paper and appendix, including code, datasets, and models. Licenses and terms of use are explicitly stated and respected.

\item {\bf New assets}
    \item[] Question: Are new assets introduced in the paper well documented and is the documentation provided alongside the assets?
    \item[] Answer: \answerYes{} 
    \item[] Justification: We release new code, datasets, and models. All new assets are thoroughly documented, including usage instructions and data format specifications, which are provided alongside the release.

\item {\bf Crowdsourcing and research with human subjects}
    \item[] Question: For crowdsourcing experiments and research with human subjects, does the paper include the full text of instructions given to participants and screenshots, if applicable, as well as details about compensation (if any)? 
    \item[] Answer: \answerNA{} 
    \item[] Justification: This paper does not involve any form of crowdsourcing or research with human subjects.

\item {\bf Institutional review board (IRB) approvals or equivalent for research with human subjects}
    \item[] Question: Does the paper describe potential risks incurred by study participants, whether such risks were disclosed to the subjects, and whether Institutional Review Board (IRB) approvals (or an equivalent approval/review based on the requirements of your country or institution) were obtained?
    \item[] Answer: \answerNA{} 
    \item[] Justification: This paper does not involve crowdsourcing or research with human subjects, so IRB approval is not applicable.

\item {\bf Declaration of LLM usage}
    \item[] Question: Does the paper describe the usage of LLMs if it is an important, original, or non-standard component of the core methods in this research? Note that if the LLM is used only for writing, editing, or formatting purposes and does not impact the core methodology, scientific rigorousness, or originality of the research, declaration is not required.
    \item[] Answer: \answerNA{} 
    \item[] Justification: This paper does not involve the use of LLMs as an important, original, or non-standard component of the core methodology.

\end{enumerate}

%% file: section/appendix.tex
\newpage

\appendix

\section*{Appendix}

\section*{Table of Contents}

\begin{itemize}
    \item Appendix \ref{app:saturn_pseudo}: Pseudocode of \method Algorithm
    \item Appendix \ref{app:sat_constructor}: Procedure of SAT Constructor
    \item Appendix \ref{app:sat_estimation}: Derivation and Validity of the SAT Difficulty Estimation
    \item Appendix \ref{app:sat_training}: Training Schedule for \method-1.5B and \method-7B
    \item Appendix \ref{app:eval_hyperparams}: Evaluation Hyperparameters
    \item Appendix \ref{app:eval_prompts}: Prompt Templates
    \item Appendix \ref{app:rq1}: Detailed Performances of \method Models on \dataset
    \item Appendix \ref{app:ablation}: Ablation Details for \method
    \item Appendix \ref{app:stronger}: Behavior of Stronger LLMs on \method Tasks
    \item Appendix \ref{app:case_study}: Examples of LLMs Reasoning Trajectories
    \item Appendix \ref{app:related_work}: Comparisons of Existing Reasoning Tasks
    \item Appendix~\ref{app:word_cloud}: \method-7B Word Cloud on GPQA Diamond
\end{itemize}

\input{section/appendix_1}
\input{section/appendix_2}

\input{section/appendix_3}

\section{Word Cloud of \method-7B’s Outputs on GPQA Diamond}
\label{app:word_cloud}

\begin{figure}[htbp]
\centering
\includegraphics[width=0.7\linewidth]{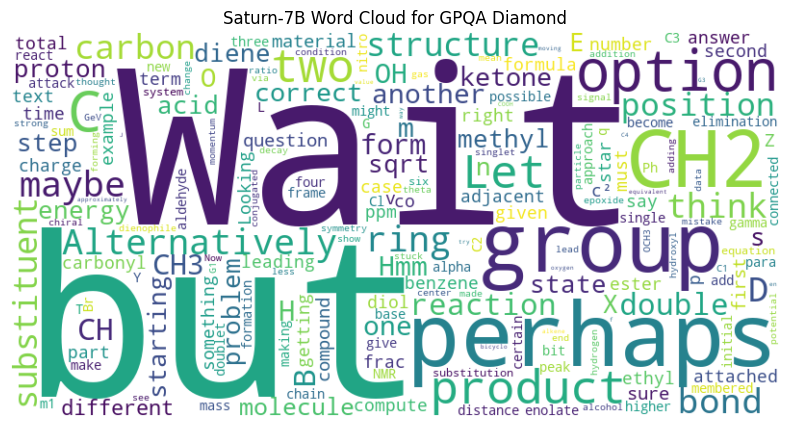}
\caption{Word cloud of \method-7B's generated answers on GPQA Diamond. Frequently used tokens are shown in larger fonts.}
\label{fig:word_cloud}
\end{figure}

Figure~\ref{fig:word_cloud} shows the word cloud of \method-7B’s generated answers on GPQA Diamond, highlighting its frequent use of self-verification patterns in reasoning.

%% file: section/appendix_1.tex
\section{Pseudocode of \method Algorithm and Hyperparameters}
\label{app:saturn_pseudo}

Algorithm~\ref{alg:curriculum} presents the complete algorithmic workflow of \method.

\input{table/algorithm_saturn}

Table~\ref{tab:saturn_hyperparams} shows all \method hyperparameters.

\begin{table}[ht]
\centering
\caption{\method Hyperparameters}
\label{tab:saturn_hyperparams}
\begin{tabular}{lcc}
\toprule
\textbf{Parameter} & \textbf{\method-1.5B} & \textbf{\method-7B} \\
\midrule
Initial $(n, k, l)$                         & (3, 5, 5)         & (3, 5, 13)       \\
\texttt{pass@k} threshold $\epsilon$        & 0.5              & 0.75              \\
Training set size per step (\texttt{Train\_size})     & 250              & 250     \\
Validation set size per step (\texttt{Val\_size})     & 40               & 40      \\
Difficulty increment (\texttt{D\_step})                & 1                & 2      \\
Curriculum iterations                       & 2                & 2                 \\
Max GRPO steps per level                    & 10               & 10                \\
\bottomrule
\end{tabular}
\end{table}


\section{Construction Procedure of SAT Instances}
\label{app:sat_constructor}

This appendix describes the implementation details of the \texttt{SAT\_Construction} algorithm introduced in Section~\ref{sec:2_1}. The goal is to generate $m$ satisfiable $(n, k, l)$-SAT instances in conjunctive normal form (CNF), each constructed to be consistent with a known Boolean solution. The algorithm ensures diversity and uniformity across sampled instances.

The construction procedure is outlined in Algorithm~\ref{alg:constructor} and consists of the following steps:

\begin{enumerate}
    \item A Boolean solution is randomly generated for the $k$ variables.
    \item The constructor randomly selects $n$ variables from the $k$ variables to form a clause.
    \item Under the constraint of satisfying the solution, the $n$ variables are randomly negated, resulting in $2^n - 1$ satisfiable clauses per variable set. From these steps, we uniformly obtain a large number of single-clause samples from the total of $2^k \cdot \binom{k}{n} \cdot (2^n - 1)$ valid $(\texttt{solution}, \texttt{clause})$ pairs, with an upper bound set to $100{,}000$.
    \item These $(\texttt{solution}, \texttt{clause})$ pairs are grouped into clusters based on the same solution.
    \item Within each cluster, we randomly select $l$ clauses and shuffle their order to construct a full SAT instance that satisfies the corresponding solution.
    \item Finally, we uniformly sample across clusters and remove duplicates to obtain a total of $m$ diverse SAT instances.
\end{enumerate}

\input{table/algorithm_constructor}


\section{Derivation and Validity of Difficulty Estimation for SAT Tasks on LLMs}
\label{app:sat_estimation}

As stated in Section~\ref{sec:2_2}, we propose a composite difficulty function Eq.~(\ref{eq:D}) to estimate the reasoning difficulty of SAT problems for LLMs. This difficulty score combines a sparsity-based estimate of solution density with a structural complexity adjustment.

\subsection*{Step 1: Sparsity-Based Estimate ($D_1$)}

We first estimate task difficulty by measuring the ratio between symbolic search cost and expected solution space size.

The symbolic search cost is approximately proportional to the number of variable-symbol combinations across decoding steps. For a problem with $k$ variables and $l$ decoding steps, we estimate:
\begin{equation}
\mathbf{Search\ Cost} \propto k \cdot l
\end{equation}

For Boolean constraint problems like SAT, the number of satisfying assignments is sparse. Assuming a random clauses, the expected number of valid solutions is:
\begin{equation}
\mathbf{Expected\ Solutions} \approx 2^n - 1 \approx 2^n
\end{equation}
where we approximate $2^n - 1$ by $2^n$ for analytical simplicity.

Taking the ratio and applying a logarithmic transform yields:
\begin{equation}
D_1(n, k, l) = \log_2(k \cdot l) - \log_2(2^n) = \log_2(k) + \log_2(l) - n
\end{equation}

\subsection*{Step 2: Structural Complexity Adjustment ($D_2$)}

In addition to solution sparsity, symbolic reasoning difficulty also depends on the structural properties of the input formula. We construct a structure-aware term $D_2(n, k, l)$ based on two contributing factors.

First, consider the symbolic reuse density: when \(k\) variables are reused across \(n\) clauses, each variable is, on average, involved in \(n/k\) constraints. This increases the interdependency between clauses. Since higher reuse leads to greater symbolic entanglement, making factorization more challenging for the model, we define the inverse ratio \(\frac{k}{n}\) as a proxy for the structural cost:

\begin{equation}
\mathbf{Variable\ Interaction\ Cost}  = \frac{k}{n}
\end{equation}

Second, the clause length $l$ determines the number of symbols that each clause contains. Longer clauses introduce more intra-clause dependencies, increasing local reasoning complexity. We approximate this with:
\begin{equation}
\mathbf{Clause\ Width\ Cost} = \log_2(l)
\end{equation}

These two components affect reasoning difficulty independently—one globally through variable sharing, and the other locally through clause complexity. We, therefore, combine them additively into:
\begin{equation}
D_2(n, k, l) = \frac{k}{n} + \log_2(l)
\end{equation}

This additive form is justified as both terms grow monotonically with symbolic complexity and are approximately aligned in scale, enabling stable and interpretable difficulty estimation.

\subsection*{Step 3: Final Composite Metric}

Combining both components yields:
\begin{align}
D(n, k, l) &= D_1(n, k, l) + D_2(n, k, l) \\
&= \log_2(k) + \log_2(l) - n + \frac{k}{n} + \log_2(l) \\
&= \log_2(k) + 2\log_2(l) - n + \frac{k}{n}
\label{eq:app12}
\end{align}

This final composite metric provides a stable and interpretable approximation of symbolic difficulty for LLMs, taking into account both sparsity and structure.

\subsection*{Step 4: Ablation Study about estimation metric}

To further validate the necessity and effectiveness of Eq. (\ref{eq:app12}) composite metric, we conduct an ablation study comparing alternative formulations. Table~\ref{tab:metric_comparison} reports the $R^2$ correlation of each metric with LLM performance (\texttt{pass@3}) across multiple model scales.

\begin{table}[h]
\centering
\caption{Metric comparison across models.}
\label{tab:metric_comparison}

\setlength{\tabcolsep}{2pt} 
\resizebox{1.0\textwidth}{!}{%
\begin{tabular}{lcccccc}
\toprule
Metric Formula & R1-Qwen-1.5B & Saturn-1.5B & R1-Qwen-7B & Saturn-7B & Avg. & Std. Dev. \\
\midrule
$-k - l \cdot \log_2(1 - \frac{1}{2^n})$ & 0.507 & 0.537 & 0.132 & 0.000 & 0.294 & 0.269 \\
$\alpha \cdot \frac{k}{n} + \beta \cdot \log_2(l), \ \alpha=2, \ \beta=1$ & 0.428 & 0.478 & 0.568 & 0.508 & 0.496 & 0.059 \\
$\alpha \cdot \frac{k}{n} + \beta \cdot \log_2(l), \ \alpha=1, \ \beta=1$ & 0.240 & 0.279 & 0.719 & 0.826 & 0.516 & 0.300 \\
$\log_2(k \cdot l) - \log_2(2^{n} - 1)$ & 0.875 & 0.893 & 0.451 & 0.157 & 0.594 & 0.356 \\
$\log_2(k) + 2 \cdot \log_2(l) - n + \frac{k}{n}$ & 0.707 & 0.746 & 0.724 & 0.501 & \textbf{0.670} & 0.113 \\
\bottomrule
\end{tabular}%
}
\end{table}

Simpler metrics that consider only sparsity or only structural complexity perform worse overall, confirming that both components are essential for accurately capturing task difficulty. Our metric $\log_2(k) + 2 \cdot \log_2(l) - n + \frac{k}{n}$ combines both solution sparsity and structural complexity, which achieves the best overall correlation. The ratio between the solution space and the LLM’s search space is the most crucial aspect. On SATURN‑7B, the $R^2$ value is about 0.5 because the LLM already achieves over 90\% accuracy on easy problems, which limits the observable linear correlation in that range. The corresponding figures are shown in Figures~\ref{fig:difficulty_vs_accuracy_small} and~\ref{fig:difficulty_vs_accuracy_big}.


\section{Training Schedule for \method-1.5B and \method-7B}
\label{app:sat_training}

This appendix provides additional details on the \method-1.5B and \method-7B training. We conduct all experiments on 8 NVIDIA A100 40GB GPUs. We use the \texttt{OpenRLHF} framework\footnote{\url{https://github.com/OpenRLHF/OpenRLHF}} for GRPO training. The framework is designed to make RL training simple and user-friendly, which works well in our experiments.  The hyperparameters used in training are summarized in Table~\ref{tab:training_hyperparams}. All other parameters not listed above consistently follow the default settings of \texttt{OpenRLHF}.

\begin{table}[ht]
\centering
\caption{\texttt{OpenRLHF} Training Hyperparameters}
\label{tab:training_hyperparams}
\begin{tabular}{lcc}
\toprule
\textbf{Parameter} & \textbf{\method-1.5B} & \textbf{\method-7B} \\
\midrule
Actor learning rate        & $5 \times 10^{-7}$ & $5 \times 10^{-7}$ \\
Initial KL coefficient     & $1 \times 10^{-3}$ & $1 \times 10^{-3}$ \\
Batch size (train)         & 2                  & 2 \\
Batch size (rollout)       & 2                  & 2 \\
Samples per prompt         & 8                  & 8 \\
Prompt length (max)        & 1024               & 1024 \\
Generation length (max)    & 10000              & 8192 \\
Temperature                & 0.8                & 1.0 \\
Zero redundancy stage      & 3                  & 3 \\
Use bf16                   & Yes                & Yes \\
KL estimator               & k3                 & k3 \\
Advantage estimator        & GroupNorm          & GroupNorm \\
Use reward normalization   & Yes                & Yes \\
\bottomrule
\end{tabular}
\end{table}

\begin{figure}[htbp]
\centering
\includegraphics[width=0.85\linewidth]{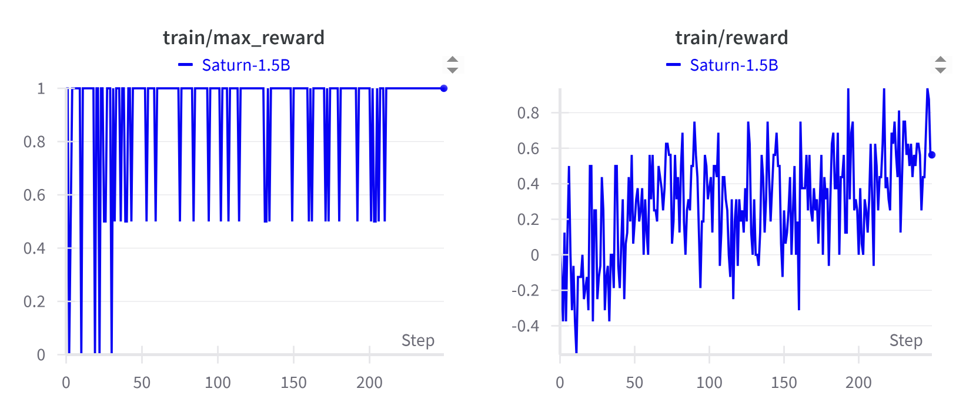}
\caption{Training curves of various metrics for \method-1.5B.}
\label{fig:Saturn_Curve_1.5B}
\end{figure}

\begin{figure}[htbp]
\centering
\includegraphics[width=0.85\linewidth]{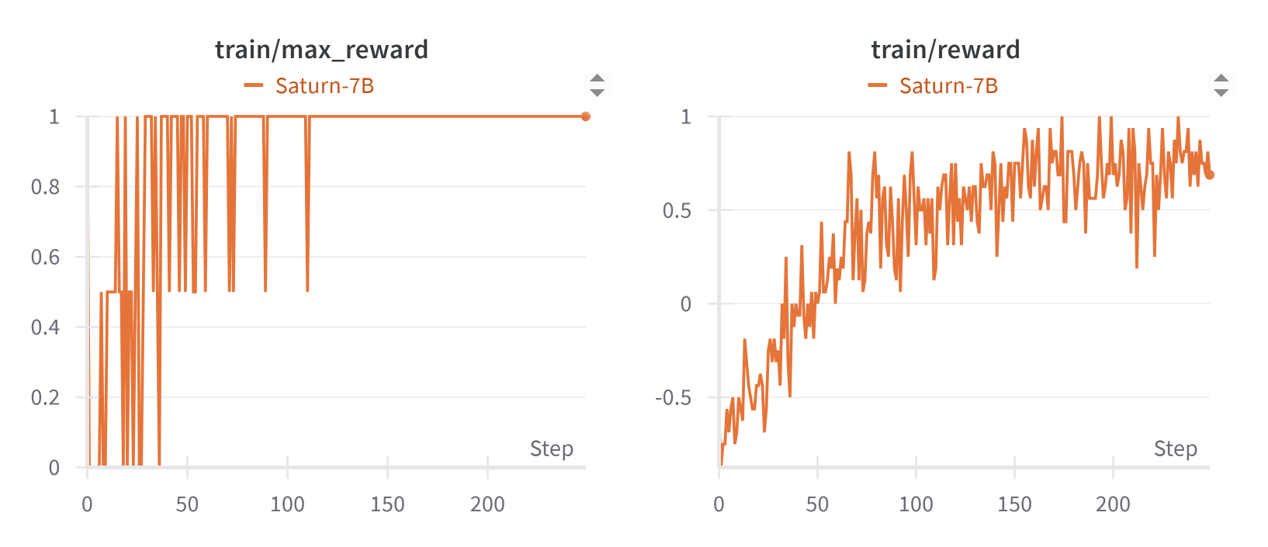}
\caption{Training curves of various metrics for \method-7B.}
\label{fig:Saturn_Curve_7B}
\end{figure}

Figure~\ref{fig:Saturn_Curve_1.5B} and Figure~\ref{fig:Saturn_Curve_7B} illustrate the training curves of \method-1.5B and \method-7B. Both models show a clear upward trend in the average reward during training, accompanied by early fluctuations that gradually stabilize. The maximum reward curves quickly reach high values and maintain them throughout most of the training process. These results indicate that both \method-1.5B and \method-7B successfully learn to generate high-reward outputs, demonstrating effective SAT reward-guided optimization.

%% file: table/algorithm_saturn.tex
\begin{algorithm}[htbp]
\caption{\texttt{\method Learning\_Loop}$(n, k, l, \pi_\theta)$  \quad \# LLM represents $\pi_\theta$ }
\label{alg:curriculum}

\begin{lstlisting}[language=Python, mathescape=true, breaklines=true]
def Increase_difficulty(n, k, l, step=1):
    """
    Increments by D_step to increase SAT difficulty.
    """
    return n, k, l + D_step

def SATURN_learning_loop(n, k, l, LLM):
    for t in range(2):  # Max total curriculum iterations
        # Step 1: Curriculum Estimation Loop
        # Generate validation set
        Val_set = SAT_Construction(n, k, l, Val_size) 
        pass_at_1 = evaluate(Val_set, LLM)  # Evaluate pass@1

        if pass_at_1 >= epsilon:
            n, k, l = Increase_difficulty(n, k, l)
        else:
            # Step 2: LLM Training Loop (only if pass_at_1 < epsilon)
            for i in range(10):  # Max GRPO steps per level
                # Generate training set
                D_train = SAT_Construction(n, k, l, Train_size)
                LLM = GRPO(LLM, D_train)

                # Re-generate validation set
                Val_set = SAT_Construction(n, k, l, Val_size)  
                pass_at_1 = evaluate(Val_set, LLM)  # Evaluate pass@1
                if pass_at_1 >= epsilon:
                    break

    return LLM
\end{lstlisting}

\end{algorithm}

%% file: table/algorithm_constructor.tex
\begin{algorithm}[htbp]
\caption{\texttt{SAT\_Construction}$(n, k, l, m)$}
\label{alg:constructor}
\begin{lstlisting}[language=python, mathescape=true, breaklines=true]
def SAT_Construction(n, k, l, m):
    V = [x_1, x_2, ..., x_k]  # k Boolean variables
    P = set()                 # pool of (solution, clause) pairs  

    # Step 1-3: Generate (solution, clause) pairs
    while len(P) < 100000: # max number of single-clause samples
        vars = sample_variables(V, n)     # select n vars from V
                                          # e.g., [x_2, x_4, ...]
        solution = random_assign(vars)    # assign 0/1 to vars
                                          # e.g., {x_2:1, x_4:0, ...}
        clause = generate(vars, solution) # create clause
                                          # e.g., {x_2 or !x_4 ...}
        P.add((solution, clause))         # store pair

    # Step 4: Group by solution
    clusters = group_by_solution(P)  # solution -> list of clauses

    # Step 5-6: Construct m SAT instances
    instances = set()
    while len(instances) < m:
        solution, clauses = sample_cluster(clusters)  # select a group
        if len(clauses) < l:
            continue
        selected_clauses = sample_clauses(clauses, l) # pick l clauses
        shuffle(selected_clauses)                     
        instances.add((solution, selected_clauses))

    return instances
\end{lstlisting}
\end{algorithm}

%% file: section/appendix_2.tex
\section{Evaluation Hyperparameters}
\label{app:eval_hyperparams}
This appendix provides additional details on the hyperparameters used in the evaluation. We use the Hugging Face \texttt{lighteval} library\footnote{\url{https://github.com/huggingface/lighteval}} for math and programming evaluations. It offers efficient benchmarks, helping us assess LLMs' performance across various tasks while ensuring both computational efficiency and high-quality results. For the evaluation of the DeepSeek-R1 series distillation models, we use \texttt{lighteval} with the \texttt{Hugging Face-Open-R1} framework\footnote{\url{https://github.com/huggingface/open-r1}}. This framework effectively reproduces the evaluation results of the R1 series models.

During the evaluation process, we follow the parameter settings from \texttt{Hugging Face-Open-R1}, as shown in Table~\ref{tab:eval_hyperparams}. Additionally, for LiveCodeBench, we also select the default \texttt{v4\_v5} version of this framework. Due to the long inference budget required by LiveCodeBench, we set the maximum response length to 16K and generate four samples per instance to estimate pass@1. All other parameters not listed and mentioned above consistently follow the default settings of \texttt{Hugging Face-Open-R1}. 

For larger closed-source models, we report the benchmark results from a public benchmark website\footnote{\url{https://artificialanalysis.ai/models}}.

\begin{table}[htbp]
\centering
\caption{Evaluation Hyperparameters for \texttt{Hugging Face-Open-R1}}
\label{tab:eval_hyperparams}
\begin{tabular}{l c}
\toprule
\textbf{Hyperparameter} & \textbf{Setting} \\
\midrule
Data type                & \texttt{bfloat16} \\
Maximum model length     & 32,768 \\
Maximum new tokens       & 32,768 \\
Temperature              & 0.6 \\
Top-$p$ (nucleus sampling) & 0.95 \\
\bottomrule
\end{tabular}
\end{table}

\section{Prompt Templates}
\label{app:eval_prompts}
\begin{figure}[htbp]
\centering
\includegraphics[width=1\linewidth]{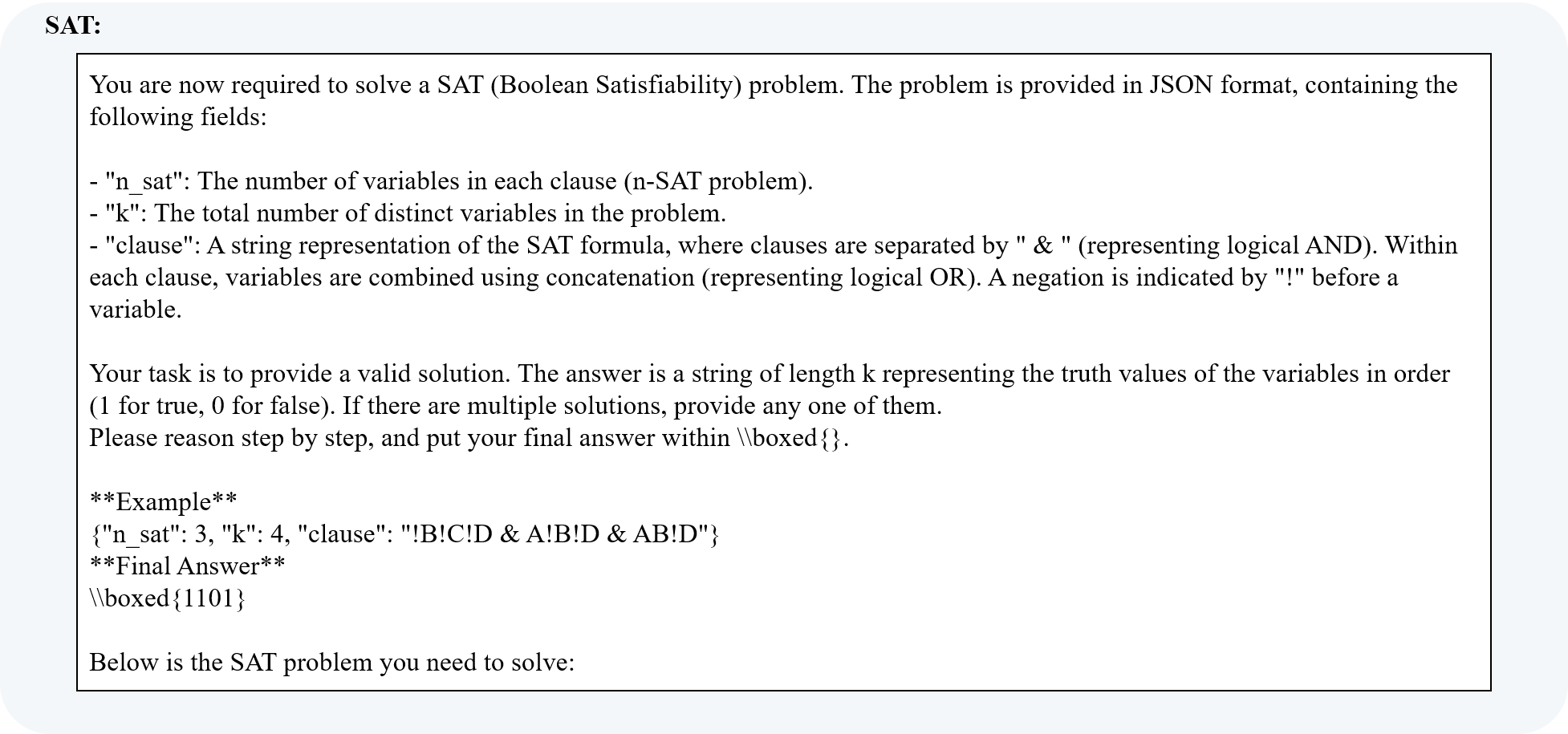}
\caption{Prompt format used for SAT problem training and evaluation.}
\label{fig:SAT_prompt}
\end{figure}

\begin{figure}[htbp]
\centering
\includegraphics[width=1\linewidth]{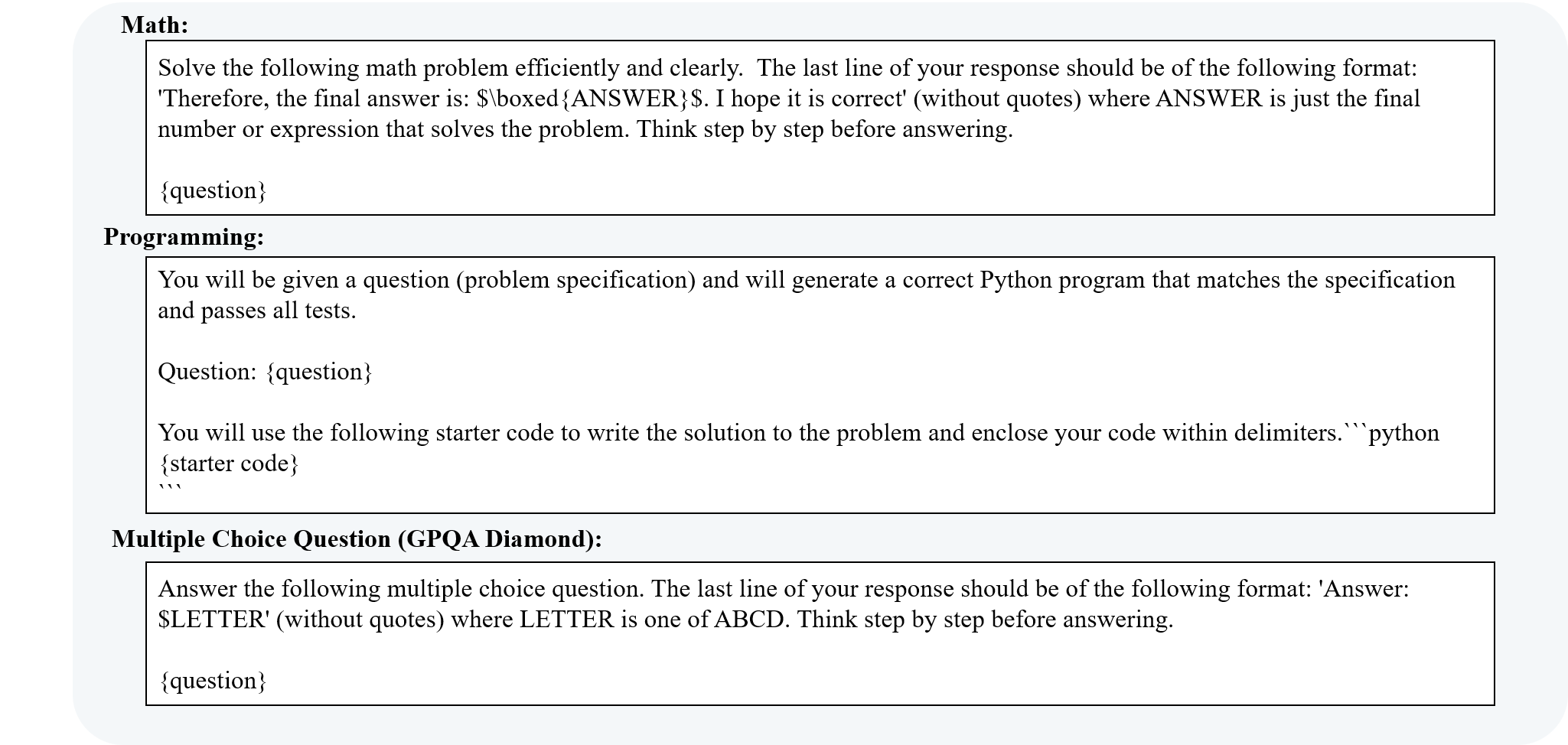}
\caption{Prompt format used for math, programming, and GPQA Diamond problems evaluation.}
\label{fig:Math_prompt}
\end{figure}

This appendix provides the prompt templates used for evaluation, ensuring consistency and reproducibility across tasks. Figure~\ref{fig:SAT_prompt} presents the format for SAT problem training and evaluation, while Figure~\ref{fig:Math_prompt} shows the template used for math, programming, and GPQA Diamond tasks.

\section{Detailed Performances of \method models on \dataset}
\label{app:rq1}

\begin{figure}[htbp]
\centering
\includegraphics[width=1\linewidth]{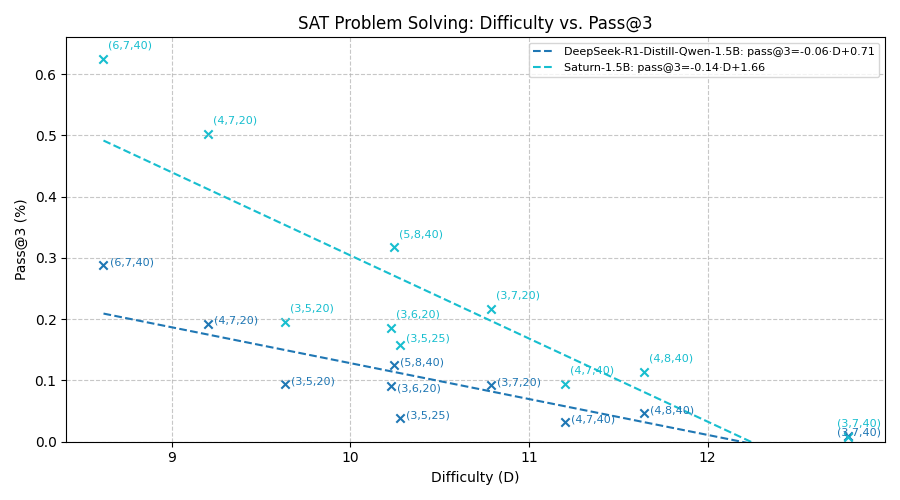}
\caption{Scatter plots of \texttt{pass@3} versus estimated difficulty $D(n, k, l)$ for DeepSeek-R1-Distill-Qwen-1.5B and \method-1.5B, with linear regression fits. The linear regression for two models achieve $ R^2 $ values of 0.707 and 0.746 respectively.}
\label{fig:difficulty_vs_accuracy_small}
\end{figure}

\begin{figure}[htbp]
\centering
\includegraphics[width=1\linewidth]{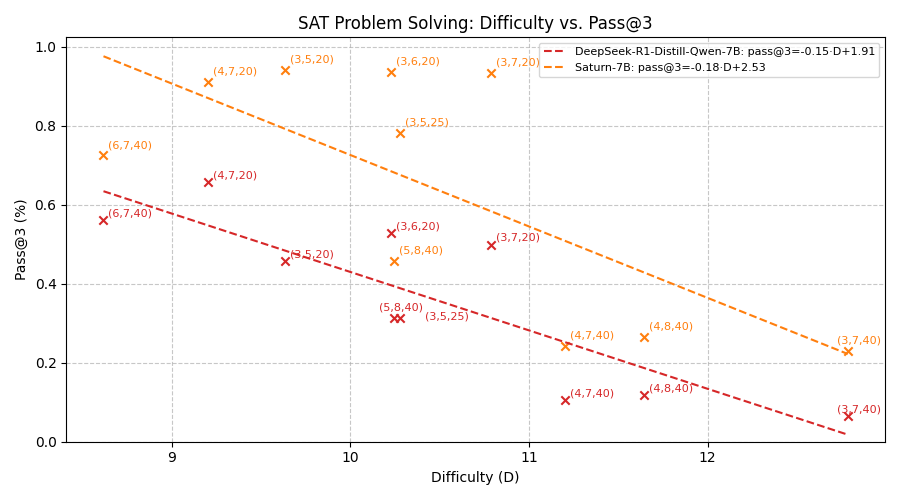}
\caption{Scatter plots of \texttt{pass@3} versus estimated difficulty $D(n, k, l)$ for DeepSeek-R1-Distill-Qwen-7B and \method-7B, with linear regression fits. The linear regression for two models achieve $ R^2 $ values of 0.724 and 0.5011 respectively.}
\label{fig:difficulty_vs_accuracy_big}
\end{figure}

This appendix provides additional details of \method-1.5B and 7B on \dataset spanning 10 harder SAT difficulty levels. Experimental results are shown in Table~\ref{app:full_sat_pass1}--\ref{app:full_sat_pass10}, and Figures~\ref{fig:difficulty_vs_accuracy_small}--\ref{fig:difficulty_vs_accuracy_big}. We summarize two key observations:

\ding{182} The pass@3 accuracy correlates strongly with the estimated SAT difficulty $D(n, k, l)$ across models. Specifically, the linear regression $R^2$ scores are 0.746 for \method-1.5B and 0.707 for DeepSeek-R1-Distill-Qwen-1.5B (Figure~\ref{fig:difficulty_vs_accuracy_small}), and 0.5011 for \method-7B and 0.724 for DeepSeek-R1-Distill-Qwen-7B (Figure~\ref{fig:difficulty_vs_accuracy_big}). These results indicate that our difficulty function effectively captures problem hardness, supporting the design of a curriculum learning schedule based on it. They also demonstrate that SAT is a reliable benchmark for evaluating reasoning capability.

\ding{183} Although our models are trained only on relatively easier SAT problems, they show consistent improvements on harder levels. As shown in Table~\ref{app:full_sat_pass1}--\ref{app:full_sat_pass10}, both \method-1.5B and \method-7B generalize well to more challenging problems, highlighting the effectiveness of our curriculum-driven training strategy.

\input{table/app_full_sat}

%% file: table/app_full_sat.tex
\begin{table}[H]
\centering
\caption{Full \texttt{pass@1} results on \dataset}
\label{app:full_sat_pass1}
\resizebox{1.0\textwidth}{!}{
\begin{tabular}{lccccccccccc}
\toprule
\textbf{Model} & \textbf{(3,7,40)} & \textbf{(3,5,25)} & \textbf{(3,5,20)} & \textbf{(3,6,20)} & \textbf{(3,7,20)} & \textbf{(4,7,40)} & \textbf{(4,8,40)} & \textbf{(4,7,20)} & \textbf{(6,7,40)} & \textbf{(5,8,40)} & \textbf{Avg.} \\
\midrule
R1-Distill-1.5B & 0.3 & 1.3 & 3.3 & 3.3 & 3.3 & 1.1 & 1.6 & 7.2 & 10.6 & 4.4 & 3.6 \\
\rowcolor[rgb]{ .741,  .843,  .933} \method-1.5B & \textbf{0.3} & \textbf{6.2} & \textbf{7.8} & \textbf{7.3} & \textbf{8.9} & \textbf{3.3} & \textbf{4.2} & \textbf{22.5} & \textbf{29.0} & \textbf{12.9} & \textbf{10.2} \\
\midrule
R1-Distill-7B & 2.3 & 12.3 & 19.2 & 23.3 & 21.5 & 3.8 & 4.5 & 30.1 & 24.4 & 12.3 & 15.4 \\
\rowcolor[rgb]{ .741,  .843,  .933} \method-7B & \textbf{8.6} & \textbf{44.7} & \textbf{66.4} & \textbf{64.5} & \textbf{64.6} & \textbf{9.2} & \textbf{10.3} & \textbf{57.3} & \textbf{36.4} & \textbf{19.2} & \textbf{38.1} \\
\bottomrule
\end{tabular}
}
\end{table}

\begin{table}[H]
\centering
\caption{Full \texttt{pass@3} results on \dataset}
\label{app:full_sat_pass3}
\resizebox{1.0\textwidth}{!}{
\begin{tabular}{lccccccccccc}
\toprule
\textbf{Model} & \textbf{(3,7,40)} & \textbf{(3,5,25)} & \textbf{(3,5,20)} & \textbf{(3,6,20)} & \textbf{(3,7,20)} & \textbf{(4,7,40)} & \textbf{(4,8,40)} & \textbf{(4,7,20)} & \textbf{(6,7,40)} & \textbf{(5,8,40)} & \textbf{Avg.} \\
\midrule
R1-Distill-1.5B & \textbf{1.0} & 3.8 & 9.4 & 9.1 & 9.3 & 3.3 & 4.7 & 19.3 & 28.9 & 12.5 & 10.1 \\
\rowcolor[rgb]{ .741,  .843,  .933} \method-1.5B & 0.8 & \textbf{15.8} & \textbf{19.5} & \textbf{18.5} & \textbf{21.6} & \textbf{9.5} & \textbf{11.3} & \textbf{50.3} & \textbf{62.5} & \textbf{31.7} & \textbf{24.2} \\
\midrule
R1-Distill-7B & 6.5 & 31.2 & 45.6 & 52.8 & 49.9 & 10.4 & 11.8 & 65.6 & 56.2 & 31.3 & 36.1 \\
\rowcolor[rgb]{ .741,  .843,  .933} \method-7B & \textbf{22.9} & \textbf{78.2} & \textbf{94.1} & \textbf{93.5} & \textbf{93.3} & \textbf{24.3} & \textbf{26.5} & \textbf{91.0} & \textbf{72.4} & \textbf{45.7} & \textbf{64.2} \\
\bottomrule
\end{tabular}
}
\end{table}

\begin{table}[H]
\centering
\caption{Full \texttt{pass@5} results on \dataset}
\label{app:full_sat_pass5}
\resizebox{1.0\textwidth}{!}{
\begin{tabular}{lccccccccccc}
\toprule
\textbf{Model} & \textbf{(3,7,40)} & \textbf{(3,5,25)} & \textbf{(3,5,20)} & \textbf{(3,6,20)} & \textbf{(3,7,20)} & \textbf{(4,7,40)} & \textbf{(4,8,40)} & \textbf{(4,7,20)} & \textbf{(6,7,40)} & \textbf{(5,8,40)} & \textbf{Avg.} \\
\midrule
R1-Distill-1.5B & \textbf{1.7} & 6.1 & 14.9 & 14.2 & 14.4 & 5.4 & 7.6 & 29.2 & 43.7 & 19.6 & 15.7 \\
\rowcolor[rgb]{ .741,  .843,  .933} \method-1.5B & 1.3 & \textbf{23.5} & \textbf{27.9} & \textbf{26.7} & \textbf{30.6} & \textbf{14.9} & \textbf{17.2} & \textbf{67.0} & \textbf{79.0} & \textbf{44.1} & \textbf{33.2} \\
\midrule
R1-Distill-7B & 10.5 & 44.7 & 62.3 & 69.2 & 66.9 & 16.0 & 17.5 & 82.9 & 74.0 & 45.4 & 48.9 \\
\rowcolor[rgb]{ .741,  .843,  .933} \method-7B & \textbf{34.4} & \textbf{89.5} & \textbf{98.5} & \textbf{98.6} & \textbf{98.2} & \textbf{35.9} & \textbf{38.6} & \textbf{97.5} & \textbf{86.4} & \textbf{62.4} & \textbf{74.0} \\
\bottomrule
\end{tabular}
}
\end{table}

\begin{table}[H]
\centering
\caption{Full \texttt{pass@7} results on \dataset}
\label{app:full_sat_pass7}
\resizebox{1.0\textwidth}{!}{
\begin{tabular}{lccccccccccc}
\toprule
\textbf{Model} & \textbf{(3,7,40)} & \textbf{(3,5,25)} & \textbf{(3,5,20)} & \textbf{(3,6,20)} & \textbf{(3,7,20)} & \textbf{(4,7,40)} & \textbf{(4,8,40)} & \textbf{(4,7,20)} & \textbf{(6,7,40)} & \textbf{(5,8,40)} & \textbf{Avg.} \\
\midrule
R1-Distill-1.5B & \textbf{2.3} & 8.1 & 19.9 & 18.9 & 18.9 & 7.6 & 10.1 & 37.4 & 55.7 & 25.8 & 20.5 \\
\rowcolor[rgb]{ .741,  .843,  .933} \method-1.5B & 1.8 & \textbf{30.1} & \textbf{34.2} & \textbf{32.8} & \textbf{37.5} & \textbf{19.8} & \textbf{22.1} & \textbf{78.2} & \textbf{87.8} & \textbf{52.5} & \textbf{39.7} \\
\midrule
R1-Distill-7B & 14.3 & 54.6 & 73.6 & 78.7 & 77.8 & 20.9 & 22.2 & 91.6 & 84.3 & 56.3 & 57.4 \\
\rowcolor[rgb]{ .741,  .843,  .933} \method-7B & \textbf{43.7} & \textbf{94.2} & \textbf{99.6} & \textbf{99.7} & \textbf{99.5} & \textbf{45.0} & \textbf{47.6} & \textbf{99.2} & \textbf{92.2} & \textbf{73.5} & \textbf{79.4} \\
\bottomrule
\end{tabular}
}
\end{table}

\begin{table}[H]
\centering
\caption{Full \texttt{pass@10} results on \dataset}
\label{app:full_sat_pass10}
\resizebox{1.0\textwidth}{!}{
\begin{tabular}{lccccccccccc}
\toprule
\textbf{Model} & \textbf{(3,7,40)} & \textbf{(3,5,25)} & \textbf{(3,5,20)} & \textbf{(3,6,20)} & \textbf{(3,7,20)} & \textbf{(4,7,40)} & \textbf{(4,8,40)} & \textbf{(4,7,20)} & \textbf{(6,7,40)} & \textbf{(5,8,40)} & \textbf{Avg.} \\
\midrule
R1-Distill-1.5B & \textbf{3.3} & 10.6 & 26.8 & 25.2 & 24.6 & 10.8 & 14.5 & 47.9 & 69.3 & 33.7 & 26.7 \\
\rowcolor[rgb]{ .741,  .843,  .933} \method-1.5B & 2.5 & \textbf{38.7} & \textbf{41.3} & \textbf{39.5} & \textbf{45.7} & \textbf{26.2} & \textbf{28.0} & \textbf{89.6} & \textbf{101.0} & \textbf{75.9} & \textbf{46.7} \\
\midrule
R1-Distill-7B & 19.6 & 65.0 & 84.7 & 86.6 & 87.8 & 27.2 & 27.9 & 97.4 & 92.8 & 68.6 & 65.8 \\
\rowcolor[rgb]{ .741,  .843,  .933} \method-7B & \textbf{54.9} & \textbf{97.1} & \textbf{99.9} & \textbf{99.9} & \textbf{100.0} & \textbf{55.0} & \textbf{57.3} & \textbf{99.8} & \textbf{95.3} & \textbf{84.7} & \textbf{84.4} \\
\bottomrule
\end{tabular}
}
\end{table}

%% file: section/appendix_3.tex
\section{Ablation Studies for \method}
\label{app:ablation}

This appendix presents the ablation studies for \method, as shown in Table~\ref{tab:ablation_1} and Table~\ref{tab:ablation_2}. Each training setting is denoted as \((n,k,l)\times \texttt{Train\_size}\), where \((n,k,l)\) controls SAT construction and \texttt{Train\_size} is the number of training examples. Here, \texttt{Train\_size} can also be written as \texttt{Train\_size} = $D \times \texttt{num}$, where $D$ is the number of difficulty levels and \texttt{num} is the number of samples per level in \dataset (Figure~\ref{fig:difficulty_vs_accuracy_big} and \ref{fig:difficulty_vs_accuracy_small}). These experiments validate the effectiveness of curriculum learning and the design of various SAT training configurations.

\input{table/ablation}

In Table~\ref{tab:ablation_1}, we evaluate the impact of SAT difficulty, training budgets, and curriculum structure. We draw two key conclusions:

\ding{182} \textbf{SATs that are too easy or too hard hinder model learning.} Training solely on easy \((3,5,10)\times500\) or hard \((3,5,15)\times500\) instances results in lower average scores (62.2 and 53.5, respectively). In contrast, moderate-difficulty SATs \((3,5,13)\times500\) yield a higher score of 62.8, showing that balanced difficulty is essential for effective reasoning development.

\ding{183} \textbf{Multi-stage curriculum learning outperforms flat or mixed training.} Curriculum learning with progressively increasing SAT difficulty \((3,5,13)\times500 + (3,5,15)\times500\) achieves the highest average score of 63.4. In contrast, one-epoch mixed training \(((3,5,13)+(3,5,15))\times500\) only reaches 61.7, despite using the same total number of examples. Furthermore, simply scaling up a single-stage setting \((3,5,13)\times1000\) yields 62.5, which is also inferior to the curriculum. These results indicate that progressive difficulty scheduling is more effective than either flat or mixed training with the same or larger data budget.

Table~\ref{tab:ablation_2} further investigates the impact of training thresholds and step sizes under a fixed total training budget.

\ding{184} \textbf{Gradual difficulty progression outperforms random shuffling of difficulty levels.} Both \((n,k,l)\times100\times10\) and \((n,k,l)\times200\times5\) perform better when difficulty levels follow a gradual progression (62.3 and 62.6), compared to random shuffling of difficulty levels (55.1 and 62.2). This demonstrates that a curriculum learning approach with progressive difficulty scheduling is more effective.

\ding{185} \textbf{Excessively fine-grained difficulty levels hinder performance.} 
Training with overly fine-grained difficulty levels, such as \((n,k,l)\times100\times10\), results in lower performance (55.1) compared to coarser steps like \((n,k,l)\times200\times5\) (62.6). Both of these configurations perform worse than the two-stage curriculum \((3,5,13)\times500 + (3,5,15)\times500\), which achieves the highest performance with an average score of 63.4. 
This indicates that excessively fine-grained difficulty levels prevent the model from effectively mastering each level before moving on to the next, hindering overall learning.




\section{Behavior of Stronger LLMs on Extended \method Tasks}
\label{app:stronger}

This appendix demonstrates the performance of stronger LLMs on more challenging SAT tasks. The experimental results are shown in Table~\ref{tab:stronger}. Even the strongest LLMs available today cannot solve complex SAT tasks as effectively as humans using simple reflection and verification search. Due to the long CoT involved, the full LLMs' outputs are provided in the supplementary material.

\input{table/app_stronger}

To provide a baseline comparison, we also tested a CDCL SAT solver \footnote{\url{https://en.wikipedia.org/wiki/Conflict-driven_clause_learning}} on the SATURN-2.6k test set. The results are as follows:

\begin{table}[h]
\centering
\caption{CDCL SAT solver performance on SATURN-2.6k test set.}
\label{tab:cdcl_performance}
\begin{tabular}{l c}
\toprule
Metric & Value \\
\midrule
Total instances & 1000 \\
Satisfiable & 1000 \\
Unsatisfiable & 0 \\
Valid SAT Models & 1000 \\
Model Accuracy & 100.00\% \\
Total time taken (s) & 0.14 \\
\bottomrule
\end{tabular}
\end{table}

Table~\ref{tab:cdcl_performance} shows that the CDCL SAT solver significantly outperforms current LLMs like DeepSeek-R1, both in terms of runtime and accuracy. \method applied to today's strongest LLMs still has great potential.

\section{Examples of Different LLMs' Reasoning Trajectories}
\label{app:case_study}

This appendix presents examples of different LLMs' reasoning trajectories, as shown in Figures~\ref{fig:rq4_case_study_402} and \ref{fig:rq4_case_study_sat}. The full LLMs' outputs are provided in the supplementary material.

\begin{figure}[htbp]
\centering
\includegraphics[width=1\linewidth]{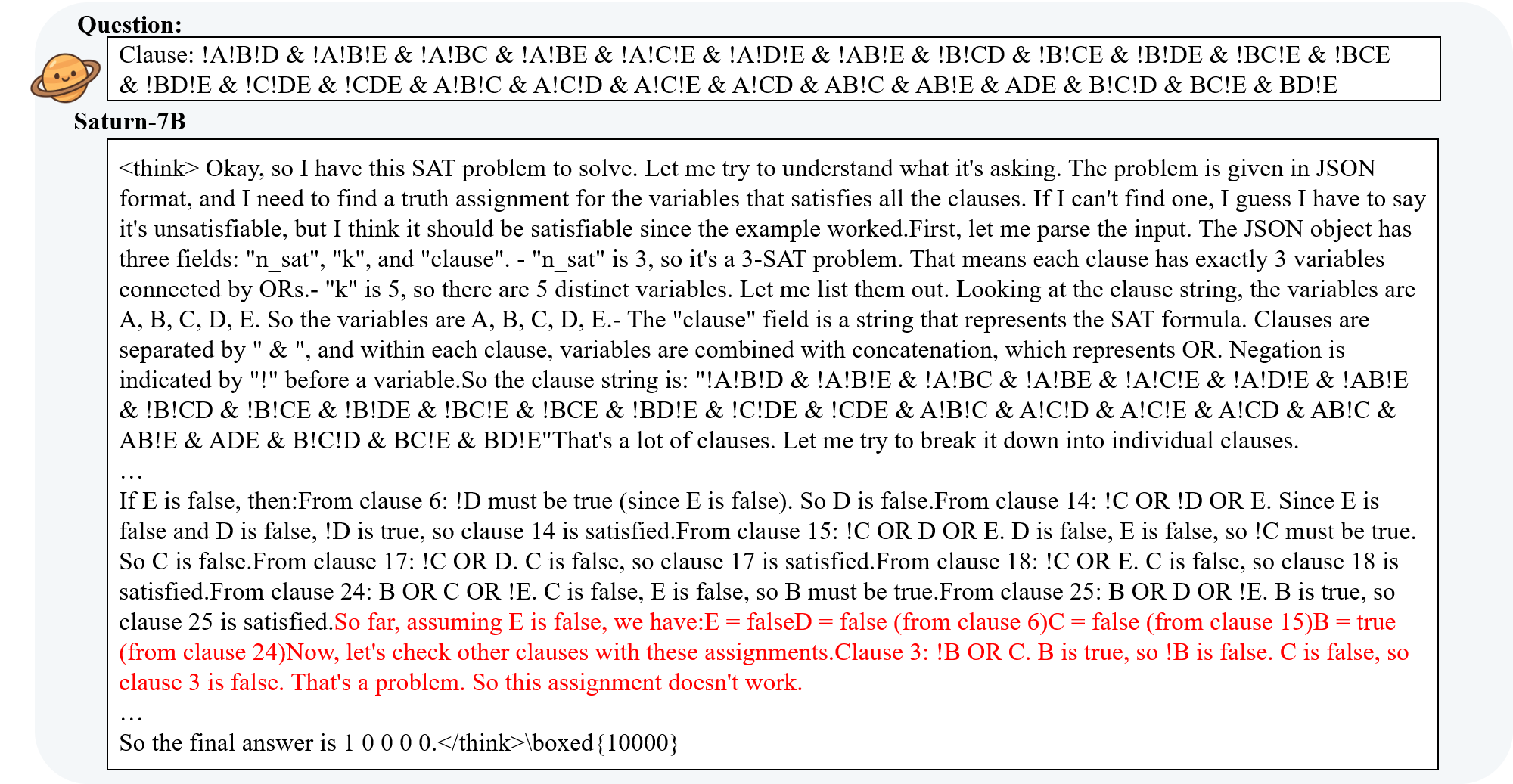}
\caption{Case study of \method-7B's reasoning trajectory on a SAT instance.}
\label{fig:rq4_case_study_sat}
\end{figure}

\begin{figure}[htbp]
\centering
\includegraphics[width=1\linewidth]{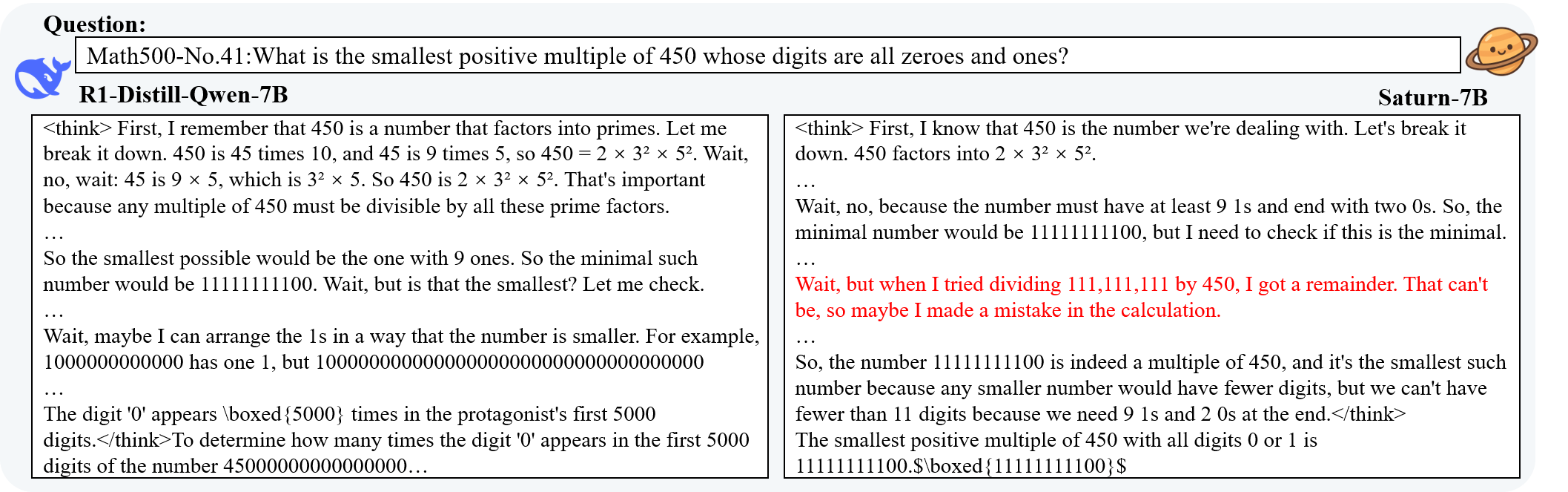}
\caption{Comparison of reasoning trajectories on Math 500 Question No.41 between two models.}
\label{fig:rq4_case_study_41}
\end{figure}

In the case of Math500-41 (Table~\ref{fig:rq4_case_study_41}), \method improves its ability to avoid unnecessary calculations and dead-end reasoning paths. By leveraging the self-verification patterns learned from SAT tasks, such as "I made a mistake earlier," \method-7B can better navigate through the problem, discarding invalid paths earlier in the process.

In the case of Math500-402 (Table~\ref{fig:rq4_case_study_402}), \method-7B demonstrates a stronger ability to self-verify intermediate conclusions in a smaller scenario (with 4 individuals). While DeepSeek-R1-Distill-Qwen-7B also tries to identify a smaller scenario, it fails to recheck the result when an inconsistency is found, instead stating, "perhaps I'm overcomplicating this." In contrast, \method-7B can identify the error and re-verify the results within this small scenario, ultimately selecting the correct solution from two possible candidates.

In conclusion, \method-7B exhibits enhanced self-verification capabilities. LLMs sometimes confidently claim that a wrong answer is correct. Solving SAT tasks inherently involves frequent and fine-grained clause verification, which trains LLMs to perform precise checking during reasoning. The self-verification patterns learned from SAT tasks help LLMs solve math problems more effectively by selecting correct solutions from multiple options. These results suggest that the self-verification mechanisms developed during SAT learning (Table~\ref{fig:rq4_case_study_sat}) generalize well to math and programming tasks, improving the LLMs' reasoning robustness and reliability.

\section{Detailed Comparison of Constructed Reasoning Tasks}
\label{app:related_work}

\begin{figure}[htbp]
\centering
\includegraphics[width=1\linewidth]{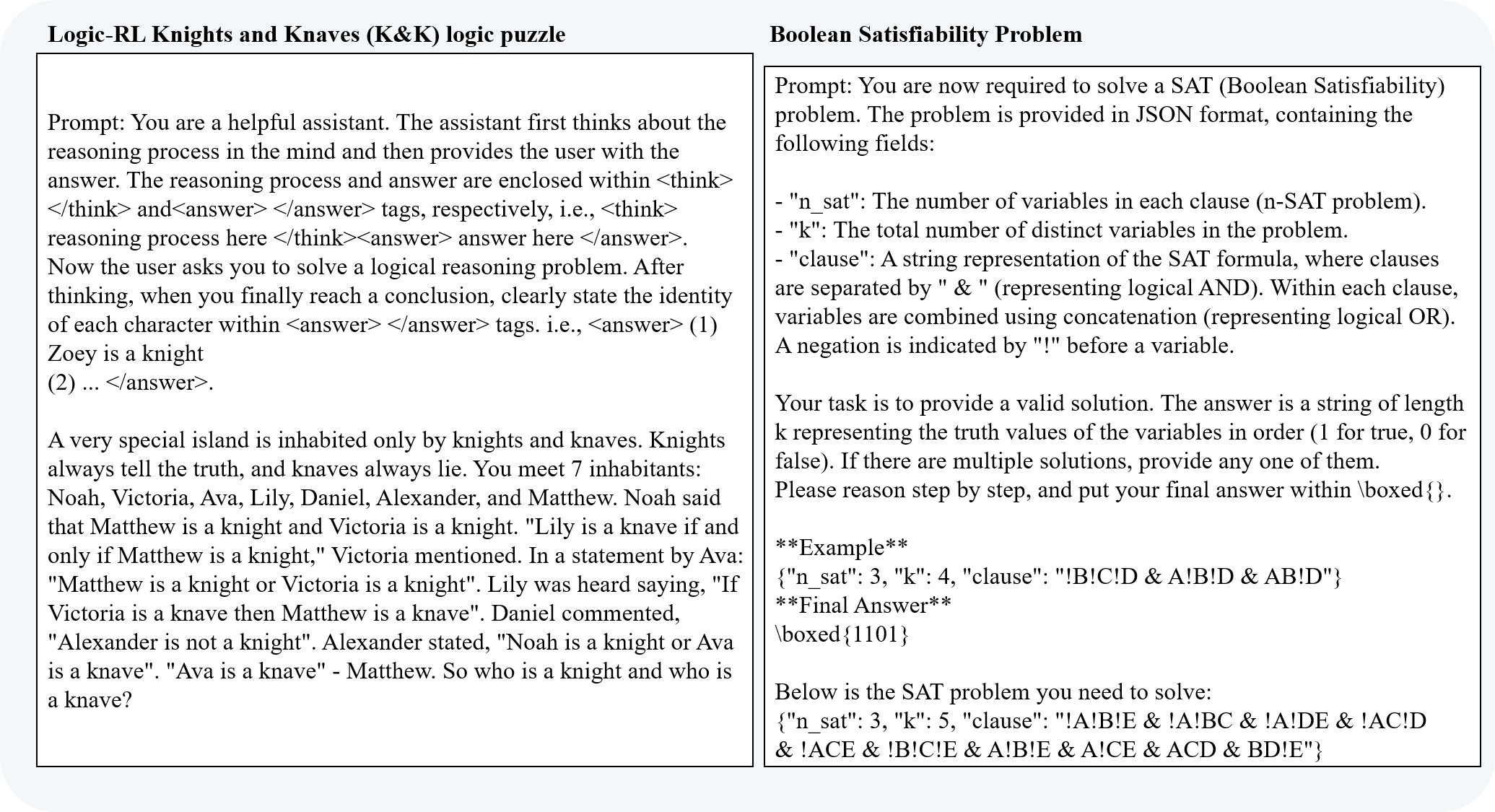}
\caption{Comparison of Knights and Knaves (K\&K) logic puzzle and SAT problem.}
\label{fig:logicRL_SAT}
\end{figure}

This appendix provides a detailed comparison between \textit{K\&K logic puzzle} and \textit{SAT problem}. Illustrative examples of each type are shown in Figure~\ref{fig:logicRL_SAT}. While both are designed to test reasoning capability, they differ in construction cost, output format complexity, and difficulty control.

\paragraph{Construction Cost.}  
K\&K puzzles require translating symbolic logic into natural language. This involves paraphrasing logical constraints into grammatically and semantically valid sentences, which increases construction cost. In contrast, SAT problems are purely symbolic and follow a standard format. As shown in \cite{sat_iclr_1}, symbolic SAT and its natural language version yield similar reasoning performance for LLMs, suggesting the symbolic form is sufficient.

\paragraph{Output Format Complexity.}  
K\&K puzzles require answers that list each character’s identity in order, such as (1) A is a knight, (2) B is a knave. This format imposes strict requirements on structure, making it harder for LLMs to follow instructions. In practice, we observe that models struggle to learn this format in early training stages. SAT problems only require a fixed-length binary string wrapped in \texttt{\textbackslash boxed\{\}}, which simplifies output and improves consistency during training.

\paragraph{Difficulty Control.}  
K\&K puzzles use the number of characters to control difficulty, which is coarse-grained. SAT problems allow fine-grained control via clause structure and variable interactions. We further define an estimation of SAT task difficulty for LLMs as \( D(n, k, l) = \log_2(k) + 2\log_2(l) - n + \frac{k}{n} \). Adding a clause to a SAT formula never decreases its difficulty, for both humans and LLMs. This makes SAT more suitable for curriculum learning.

In summary, while K\&K puzzles provide linguistic diversity, SAT problems are more efficient in construction, output consistency, and difficulty regulation, making them preferable for training LLMs at scale.

%% file: table/ablation.tex
\begin{table}[H]
    \centering
    \caption{Ablation comparison on math and programming benchmarks}
    \label{tab:ablation_1}
    \resizebox{\textwidth}{!}{
    \begin{tabular}{lcccccc}
    \toprule
    \textbf{Training Setting} & \textbf{AIME 24/25} & \textbf{AMC 22/23} & \textbf{Math500} & \textbf{GPQA-D} & \textbf{LiveCodeBench} & \textbf{Avg.} \\
    \midrule
    $(3,5,10) \times 500$          & 45.0 & 84.3 & 93.2 & \textbf{53.5}  & 35.1 & 62.2 \\
    $(3,5,13) \times 500$          & 50.0 & 83.1 & 94.6 & 50.0  & 36.1 & 62.8 \\
    $(3,5,15) \times 500$         & 35.0 & 68.7 & 86.6 & 46.0  & 31.3 & 53.5 \\
    $(3,5,13) \times 1000$         & \textbf{51.7} & 81.9 & 94.0 & 47.5  & 37.2 & 62.5 \\
    $((3,5,13)+(3,5,15)) \times 500$ (one epoch) & 43.3 & \textbf{86.7} & 93.0 & 49.5  & 35.8 & 61.7 \\
    \rowcolor[rgb]{ .741,  .843,  .933} $(3,5,13)\times500+(3,5,15)\times500$ & 48.3 & 85.5 & \textbf{95.0} & 50.5 & \textbf{37.7} & \textbf{63.4} \\
    \bottomrule
    \end{tabular}}
\end{table}

\begin{table}[H]
    \centering
    \caption{Ablation Study with Different Sampling Strategies and Training Budgets}
    \label{tab:ablation_2}
    \resizebox{\textwidth}{!}{
    \begin{tabular}{lcccccc}
    \toprule
    \textbf{Training Setting} & \textbf{AIME 24/25} & \textbf{AMC 22/23} & \textbf{Math500} & \textbf{GPQA-D} & \textbf{LiveCodeBench} & \textbf{Avg.} \\
    \midrule
    $(n,k,l)\times100\times10$ + shuffle        & 38.3 & 66.3 & 90.6 & 44.9 & 35.2 & 55.1 \\
    $(n,k,l)\times100\times10$                  & 46.7 & 85.5 & 93.2 & 50.5 & 36.9 & 62.3 \\
    $(n,k,l)\times200\times5$ + shuffle         & 48.3 & 81.9 & 93.0 & \textbf{52.0} & 35.8 & 62.2 \\
    $(n,k,l)\times200\times5$                   & 46.7 & \textbf{88.0} & 93.2 & 48.0 & 35.8 & 62.6 \\
    \rowcolor[rgb]{ .741,  .843,  .933} $(3,5,13)\times500 + (3,5,15)\times500$     & \textbf{48.3} & 85.5 & \textbf{95.0} & 50.5 & \textbf{37.7} & \textbf{63.4} \\
    \bottomrule
    \end{tabular}}
\end{table}

%% file: table/app_stronger.tex
\begin{table}[H]
    \centering
    \caption{One-shot performance of stronger LLMs on extended \method tasks. 
    Models with \ding{52} successfully solve the corresponding difficulty of SAT tasks. 
    Kimi-1.5 solves the task but with significantly longer reasoning chains.}
    \label{tab:stronger}
    \begin{tabular}{cccccc}
        \toprule
        \textbf{SAT Task (n, k, length)} & \textbf{GPT-4o} & \textbf{O1-mini} & \textbf{DeepSeek-V3} & \textbf{R1} & \textbf{Kimi-1.5} \\
        \midrule
        (3, 5, 30) & \ding{55} & \ding{52} & \ding{55} & \ding{52} & \ding{52}\textnormal{*} \\
        (4, 7, 80) & \ding{55} & \ding{55} & \ding{55} & \ding{55} & \ding{55} \\
        \bottomrule
    \end{tabular}
\end{table}

%% file: ref.bib
@article{aixcoderv2,
  author       = {Jia Li and
                  Hao Zhu and
                  Huanyu Liu and
                  Xianjie Shi and
                  He Zong and
                  Yihong Dong and
                  Kechi Zhang and
                  Siyuan Jiang and
                  Zhi Jin and
                  Ge Li},
  title        = {aiXcoder-7B-v2: Training LLMs to Fully Utilize the Long Context in
                  Repository-level Code Completion},
  journal      = {CoRR},
  volume       = {abs/2503.15301},
  year         = {2025}
}

@article{aixcoder,
  author       = {Siyuan Jiang and
                  Jia Li and
                  He Zong and
                  Huanyu Liu and
                  Hao Zhu and
                  Shukai Hu and
                  Erlu Li and
                  Jiazheng Ding and
                  Yu Han and
                  Wei Ning and
                  Gen Wang and
                  Yihong Dong and
                  Kechi Zhang and
                  Ge Li},
  title        = {aiXcoder-7B: {A} Lightweight and Effective Large Language Model for
                  Code Completion},
  journal      = {CoRR},
  volume       = {abs/2410.13187},
  year         = {2024}
}

@inproceedings{logic3,
  author       = {Mark Kaminski and
                  Tobias Tebbi},
  editor       = {Maria Paola Bonacina},
  title        = {InKreSAT: Modal Reasoning via Incremental Reduction to {SAT}},
  booktitle    = {Automated Deduction - {CADE-24} - 24th International Conference on
                  Automated Deduction, Lake Placid, NY, USA, June 9-14, 2013. Proceedings},
  series       = {Lecture Notes in Computer Science},
  volume       = {7898},
  pages        = {436--442},
  publisher    = {Springer},
  year         = {2013},
  url          = {https://doi.org/10.1007/978-3-642-38574-2\_31},
  doi          = {10.1007/978-3-642-38574-2\_31},
  bibsource    = {dblp computer science bibliography, https://dblp.org}
}

@article{logic1,
  author       = {Enrico Giunchiglia and
                  Armando Tacchella and
                  Fausto Giunchiglia},
  title        = {SAT-Based Decision Procedures for Classical Modal Logics},
  journal      = {J. Autom. Reason.},
  volume       = {28},
  number       = {2},
  pages        = {143--171},
  year         = {2002},
  url          = {https://doi.org/10.1023/A:1015071400913},
  doi          = {10.1023/A:1015071400913},
  bibsource    = {dblp computer science bibliography, https://dblp.org}
}

@inproceedings{logic2,
  author       = {Roberto Sebastiani and
                  Adolfo Villafiorita},
  editor       = {Fausto Giunchiglia},
  title        = {SAT-Based Decision Procedures for Normal Modal Logics: {A} Theoretical
                  Framework},
  booktitle    = {Artificial Intelligence: Methodology, Systems, and Applications, 8th
                  International Conference, {AIMSA} '98, Sozopol, Bulgaria, September
                  21-13, 1998, Proceedings},
  series       = {Lecture Notes in Computer Science},
  volume       = {1480},
  pages        = {377--388},
  publisher    = {Springer},
  year         = {1998},
  url          = {https://doi.org/10.1007/BFb0057460},
  doi          = {10.1007/BFB0057460},
  bibsource    = {dblp computer science bibliography, https://dblp.org}
}

@inproceedings{forgetting1,
  author       = {Zaheer Abbas and
                  Rosie Zhao and
                  Joseph Modayil and
                  Adam White and
                  Marlos C. Machado},
  title        = {Loss of Plasticity in Continual Deep Reinforcement Learning},
  booktitle    = {CoLLAs},
  series       = {Proceedings of Machine Learning Research},
  volume       = {232},
  pages        = {620--636},
  publisher    = {{PMLR}},
  year         = {2023}
}

@article{forgetting2,
  author       = {Shibhansh Dohare and
                  J. Fernando Hernandez{-}Garcia and
                  Qingfeng Lan and
                  Parash Rahman and
                  A. Rupam Mahmood and
                  Richard S. Sutton},
  title        = {Loss of plasticity in deep continual learning},
  journal      = {Nat.},
  volume       = {632},
  number       = {8026},
  pages        = {768--774},
  year         = {2024}
}

@misc{aime,
    year = {2024--2025},
    title={AIME Datasets},
    url = {https://artofproblemsolving.com/wiki/index.php/AIME_Problems_and_Solutions},
}

@misc{amc,
  year = {2022--2023},
  title = {AMC Datasets},
  url = {https://artofproblemsolving.com/wiki/index.php/AMC_12_Problems_and_Solutions},
}

@article{Meta-Abilities,
  author       = {Zhiyuan Hu and
                  Yibo Wang and
                  Hanze Dong and
                  Yuhui Xu and
                  Amrita Saha and
                  Caiming Xiong and
                  Bryan Hooi and
                  Junnan Li},
  title        = {Beyond 'Aha!': Toward Systematic Meta-Abilities Alignment
                  in Large Reasoning Models},
  journal      = {CoRR},
  volume       = {abs/2505.10554},
  year         = {2025}
}

@article{Cognitive,
  author       = {Kanishk Gandhi and
                  Ayush Chakravarthy and
                  Anikait Singh and
                  Nathan Lile and
                  Noah D. Goodman},
  title        = {Cognitive Behaviors that Enable Self-Improving Reasoners, or, Four
                  Habits of Highly Effective STaRs},
  journal      = {CoRR},
  volume       = {abs/2503.01307},
  year         = {2025}
}

@article{GPQA,
  author       = {David Rein and
                  Betty Li Hou and
                  Asa Cooper Stickland and
                  Jackson Petty and
                  Richard Yuanzhe Pang and
                  Julien Dirani and
                  Julian Michael and
                  Samuel R. Bowman},
  title        = {{GPQA:} {A} Graduate-Level Google-Proof Q{\&}A Benchmark},
  journal      = {CoRR},
  volume       = {abs/2311.12022},
  year         = {2023}
}

@inproceedings{alignment_tax,
  author       = {Long Ouyang and
                  Jeffrey Wu and
                  Xu Jiang and
                  Diogo Almeida and
                  Carroll L. Wainwright and
                  Pamela Mishkin and
                  Chong Zhang and
                  Sandhini Agarwal and
                  Katarina Slama and
                  Alex Ray and
                  John Schulman and
                  Jacob Hilton and
                  Fraser Kelton and
                  Luke Miller and
                  Maddie Simens and
                  Amanda Askell and
                  Peter Welinder and
                  Paul F. Christiano and
                  Jan Leike and
                  Ryan Lowe},
  title        = {Training language models to follow instructions with human feedback},
  booktitle    = {NeurIPS},
  year         = {2022}
}

@article{SFT-RL,
  author       = {Tianzhe Chu and
                  Yuexiang Zhai and
                  Jihan Yang and
                  Shengbang Tong and
                  Saining Xie and
                  Dale Schuurmans and
                  Quoc V. Le and
                  Sergey Levine and
                  Yi Ma},
  title        = {{SFT} Memorizes, {RL} Generalizes: {A} Comparative Study of Foundation
                  Model Post-training},
  journal      = {CoRR},
  volume       = {abs/2501.17161},
  year         = {2025}
}

@article{lcb,
  author       = {Naman Jain and
                  King Han and
                  Alex Gu and
                  Wen{-}Ding Li and
                  Fanjia Yan and
                  Tianjun Zhang and
                  Sida Wang and
                  Armando Solar{-}Lezama and
                  Koushik Sen and
                  Ion Stoica},
  title        = {LiveCodeBench: Holistic and Contamination Free Evaluation of Large
                  Language Models for Code},
  journal      = {CoRR},
  volume       = {abs/2403.07974},
  year         = {2024}
}

@inproceedings{math500,
  author       = {Dan Hendrycks and
                  Collin Burns and
                  Saurav Kadavath and
                  Akul Arora and
                  Steven Basart and
                  Eric Tang and
                  Dawn Song and
                  Jacob Steinhardt},
  title        = {Measuring Mathematical Problem Solving With the {MATH} Dataset},
  booktitle    = {NeurIPS Datasets and Benchmarks},
  year         = {2021}
}

@article{z1,
  title={Z1: Efficient Test-time Scaling with Code},
  author={Yu, Zhaojian and Wu, Yinghao and Zhao, Yilun and Cohan, Arman and Zhang, Xiao-Ping},
  journal={arXiv preprint arXiv:2504.00810},
  year={2025}
}

@inproceedings{20_Question,
  author       = {Yizhe Zhang and
                  Jiarui Lu and
                  Navdeep Jaitly},
  title        = {Probing the Multi-turn Planning Capabilities of LLMs via 20 Question
                  Games},
  booktitle    = {{ACL} {(1)}},
  pages        = {1495--1516},
  publisher    = {Association for Computational Linguistics},
  year         = {2024}
}

@article{sat_blog,
  author       = {Raffaele Marino},
  title        = {Fast Analysis of the OpenAI O1-Preview Model in Solving Random {K-SAT}
                  Problem: Does the {LLM} Solve the Problem Itself or Call an External
                  {SAT} Solver?},
  journal      = {CoRR},
  volume       = {abs/2409.11232},
  year         = {2024}
}

@article{sat_iclr_1,
  author       = {Rishi Hazra and
                  Gabriele Venturato and
                  Pedro Zuidberg Dos Martires and
                  Luc De Raedt},
  title        = {Can Large Language Models Reason? {A} Characterization via 3-SAT},
  journal      = {CoRR},
  volume       = {abs/2408.07215},
  year         = {2024}
}

@article{sat_iclr_solver,
  author       = {Leyan Pan and
                  Vijay Ganesh and
                  Jacob D. Abernethy and
                  Chris Esposo and
                  Wenke Lee},
  title        = {Can Transformers Reason Logically? {A} Study in {SAT} Solving},
  journal      = {CoRR},
  volume       = {abs/2410.07432},
  year         = {2024}
}

@article{GRPO,
  author       = {Zhihong Shao and
                  Peiyi Wang and
                  Qihao Zhu and
                  Runxin Xu and
                  Junxiao Song and
                  Mingchuan Zhang and
                  Y. K. Li and
                  Y. Wu and
                  Daya Guo},
  title        = {DeepSeekMath: Pushing the Limits of Mathematical Reasoning in Open
                  Language Models},
  journal      = {CoRR},
  volume       = {abs/2402.03300},
  year         = {2024},
  url          = {https://doi.org/10.48550/arXiv.2402.03300},
  doi          = {10.48550/ARXIV.2402.03300},
  eprinttype    = {arXiv},
  eprint       = {2402.03300},
  bibsource    = {dblp computer science bibliography, https://dblp.org}
}

@article{SAT-RSB,
  author       = {Haijun Zhou},
  title        = {Long Range Frustrations in a Spin Glass Model of the Vertex Cover
                  Problem},
  journal      = {CoRR},
  volume       = {cond-mat/0411077},
  year         = {2004}
}

@inproceedings{reduction,
  author       = {Richard M. Karp},
  title        = {Reducibility Among Combinatorial Problems},
  booktitle    = {Complexity of Computer Computations},
  series       = {The {IBM} Research Symposia Series},
  pages        = {85--103},
  publisher    = {Plenum Press, New York},
  year         = {1972}
}

@inproceedings{SAT,
  author       = {Stephen A. Cook},
  title        = {The Complexity of Theorem-Proving Procedures},
  booktitle    = {{STOC}},
  pages        = {151--158},
  publisher    = {{ACM}},
  year         = {1971}
}

@article{aplhacode,
  author       = {Yujia Li and
                  David H. Choi and
                  Junyoung Chung and
                  Nate Kushman and
                  Julian Schrittwieser and
                  R{\'{e}}mi Leblond and
                  Tom Eccles and
                  James Keeling and
                  Felix Gimeno and
                  Agustin Dal Lago and
                  Thomas Hubert and
                  Peter Choy and
                  Cyprien de Masson d'Autume and
                  Igor Babuschkin and
                  Xinyun Chen and
                  Po{-}Sen Huang and
                  Johannes Welbl and
                  Sven Gowal and
                  Alexey Cherepanov and
                  James Molloy and
                  Daniel J. Mankowitz and
                  Esme Sutherland Robson and
                  Pushmeet Kohli and
                  Nando de Freitas and
                  Koray Kavukcuoglu and
                  Oriol Vinyals},
  title        = {Competition-Level Code Generation with AlphaCode},
  journal      = {CoRR},
  volume       = {abs/2203.07814},
  year         = {2022}
}

@article{kk2,
  author       = {Chulin Xie and
                  Yangsibo Huang and
                  Chiyuan Zhang and
                  Da Yu and
                  Xinyun Chen and
                  Bill Yuchen Lin and
                  Bo Li and
                  Badih Ghazi and
                  Ravi Kumar},
  title        = {On Memorization of Large Language Models in Logical Reasoning},
  journal      = {CoRR},
  volume       = {abs/2410.23123},
  year         = {2024}
}

@article{wolf2,
  author       = {Yuzhuang Xu and
                  Shuo Wang and
                  Peng Li and
                  Fuwen Luo and
                  Xiaolong Wang and
                  Weidong Liu and
                  Yang Liu},
  title        = {Exploring Large Language Models for Communication Games: An Empirical
                  Study on Werewolf},
  journal      = {CoRR},
  volume       = {abs/2309.04658},
  year         = {2023}
}

@article{codedpo,
  author       = {Kechi Zhang and
                  Ge Li and
                  Yihong Dong and
                  Jingjing Xu and
                  Jun Zhang and
                  Jing Su and
                  Yongfei Liu and
                  Zhi Jin},
  title        = {CodeDPO: Aligning Code Models with Self Generated and Verified Source
                  Code},
  journal      = {CoRR},
  volume       = {abs/2410.05605},
  year         = {2024}
}

@article{mbpp,
  author       = {Jacob Austin and
                  Augustus Odena and
                  Maxwell I. Nye and
                  Maarten Bosma and
                  Henryk Michalewski and
                  David Dohan and
                  Ellen Jiang and
                  Carrie J. Cai and
                  Michael Terry and
                  Quoc V. Le and
                  Charles Sutton},
  title        = {Program Synthesis with Large Language Models},
  journal      = {CoRR},
  volume       = {abs/2108.07732},
  year         = {2021}
}

@article{humaneval,
  author       = {Mark Chen and
                  Jerry Tworek and
                  Heewoo Jun and
                  Qiming Yuan and
                  Henrique Pond{\'{e}} de Oliveira Pinto and
                  Jared Kaplan and
                  Harri Edwards and
                  Yuri Burda and
                  Nicholas Joseph and
                  Greg Brockman and
                  Alex Ray and
                  Raul Puri and
                  Gretchen Krueger and
                  Michael Petrov and
                  Heidy Khlaaf and
                  Girish Sastry and
                  Pamela Mishkin and
                  Brooke Chan and
                  Scott Gray and
                  Nick Ryder and
                  Mikhail Pavlov and
                  Alethea Power and
                  Lukasz Kaiser and
                  Mohammad Bavarian and
                  Clemens Winter and
                  Philippe Tillet and
                  Felipe Petroski Such and
                  Dave Cummings and
                  Matthias Plappert and
                  Fotios Chantzis and
                  Elizabeth Barnes and
                  Ariel Herbert{-}Voss and
                  William Hebgen Guss and
                  Alex Nichol and
                  Alex Paino and
                  Nikolas Tezak and
                  Jie Tang and
                  Igor Babuschkin and
                  Suchir Balaji and
                  Shantanu Jain and
                  William Saunders and
                  Christopher Hesse and
                  Andrew N. Carr and
                  Jan Leike and
                  Joshua Achiam and
                  Vedant Misra and
                  Evan Morikawa and
                  Alec Radford and
                  Matthew Knight and
                  Miles Brundage and
                  Mira Murati and
                  Katie Mayer and
                  Peter Welinder and
                  Bob McGrew and
                  Dario Amodei and
                  Sam McCandlish and
                  Ilya Sutskever and
                  Wojciech Zaremba},
  title        = {Evaluating Large Language Models Trained on Code},
  journal      = {CoRR},
  volume       = {abs/2107.03374},
  year         = {2021}
}

@article{Curriculum_Learning,
  author       = {Xin Wang and
                  Yudong Chen and
                  Wenwu Zhu},
  title        = {A Survey on Curriculum Learning},
  journal      = {{IEEE} Trans. Pattern Anal. Mach. Intell.},
  volume       = {44},
  number       = {9},
  pages        = {4555--4576},
  year         = {2022}
}

@inproceedings{Verify_Step,
  author       = {Hunter Lightman and
                  Vineet Kosaraju and
                  Yuri Burda and
                  Harrison Edwards and
                  Bowen Baker and
                  Teddy Lee and
                  Jan Leike and
                  John Schulman and
                  Ilya Sutskever and
                  Karl Cobbe},
  title        = {Let's Verify Step by Step},
  booktitle    = {{ICLR}},
  publisher    = {OpenReview.net},
  year         = {2024}
}

@article{gsm8k,
  author       = {Karl Cobbe and
                  Vineet Kosaraju and
                  Mohammad Bavarian and
                  Mark Chen and
                  Heewoo Jun and
                  Lukasz Kaiser and
                  Matthias Plappert and
                  Jerry Tworek and
                  Jacob Hilton and
                  Reiichiro Nakano and
                  Christopher Hesse and
                  John Schulman},
  title        = {Training Verifiers to Solve Math Word Problems},
  journal      = {CoRR},
  volume       = {abs/2110.14168},
  year         = {2021}
}

@article{lmrl,
  author       = {Marwa Abdulhai and
                  Isadora White and
                  Charlie Snell and
                  Charles Sun and
                  Joey Hong and
                  Yuexiang Zhai and
                  Kelvin Xu and
                  Sergey Levine},
  title        = {{LMRL} Gym: Benchmarks for Multi-Turn Reinforcement Learning with
                  Language Models},
  journal      = {CoRR},
  volume       = {abs/2311.18232},
  year         = {2023}
}

@inproceedings{difficulty_controllable,
  author       = {Yifan Gao and
                  Lidong Bing and
                  Wang Chen and
                  Michael R. Lyu and
                  Irwin King},
  title        = {Difficulty Controllable Generation of Reading Comprehension Questions},
  booktitle    = {{IJCAI}},
  pages        = {4968--4974},
  publisher    = {ijcai.org},
  year         = {2019}
}

@article{wolf1,
  author       = {Shuang Wu and
                  Liwen Zhu and
                  Tao Yang and
                  Shiwei Xu and
                  Qiang Fu and
                  Yang Wei and
                  Haobo Fu},
  title        = {Enhance Reasoning for Large Language Models in the Game Werewolf},
  journal      = {CoRR},
  volume       = {abs/2402.02330},
  year         = {2024}
}

@article{puzzbench,
  author       = {Yongqi Tong and
                  Sizhe Wang and
                  Dawei Li and
                  Yifan Wang and
                  Simeng Han and
                  Zi Lin and
                  Chengsong Huang and
                  Jiaxin Huang and
                  Jingbo Shang},
  title        = {Optimizing Language Model's Reasoning Abilities with Weak Supervision},
  journal      = {CoRR},
  volume       = {abs/2405.04086},
  year         = {2024}
}

@inproceedings{self_verification,
  author       = {Yixuan Weng and
                  Minjun Zhu and
                  Fei Xia and
                  Bin Li and
                  Shizhu He and
                  Shengping Liu and
                  Bin Sun and
                  Kang Liu and
                  Jun Zhao},
  title        = {Large Language Models are Better Reasoners with Self-Verification},
  booktitle    = {{EMNLP} (Findings)},
  pages        = {2550--2575},
  publisher    = {Association for Computational Linguistics},
  year         = {2023}
}

@article{question_synthesis,
  author       = {Yuyang Ding and
                  Xinyu Shi and
                  Xiaobo Liang and
                  Juntao Li and
                  Qiaoming Zhu and
                  Min Zhang},
  title        = {Unleashing Reasoning Capability of LLMs via Scalable Question Synthesis
                  from Scratch},
  journal      = {CoRR},
  volume       = {abs/2410.18693},
  year         = {2024}
}

@article{s1,
  author       = {Niklas Muennighoff and
                  Zitong Yang and
                  Weijia Shi and
                  Xiang Lisa Li and
                  Li Fei{-}Fei and
                  Hannaneh Hajishirzi and
                  Luke Zettlemoyer and
                  Percy Liang and
                  Emmanuel J. Cand{\`{e}}s and
                  Tatsunori Hashimoto},
  title        = {s1: Simple test-time scaling},
  journal      = {CoRR},
  volume       = {abs/2501.19393},
  year         = {2025}
}

@inproceedings{stepback,
  author       = {Huaixiu Steven Zheng and
                  Swaroop Mishra and
                  Xinyun Chen and
                  Heng{-}Tze Cheng and
                  Ed H. Chi and
                  Quoc V. Le and
                  Denny Zhou},
  title        = {Take a Step Back: Evoking Reasoning via Abstraction in Large Language
                  Models},
  booktitle    = {{ICLR}},
  publisher    = {OpenReview.net},
  year         = {2024}
}

@inproceedings{cot,
  author       = {Jason Wei and
                  Xuezhi Wang and
                  Dale Schuurmans and
                  Maarten Bosma and
                  Brian Ichter and
                  Fei Xia and
                  Ed H. Chi and
                  Quoc V. Le and
                  Denny Zhou},
  title        = {Chain-of-Thought Prompting Elicits Reasoning in Large Language Models},
  booktitle    = {NeurIPS},
  year         = {2022}
}

@article{openai_o1,
  author       = {Aaron Jaech and
                  Adam Kalai and
                  Adam Lerer and
                  Adam Richardson and
                  Ahmed El{-}Kishky and
                  Aiden Low and
                  Alec Helyar and
                  Aleksander Madry and
                  Alex Beutel and
                  Alex Carney and
                  Alex Iftimie and
                  Alex Karpenko and
                  Alex Tachard Passos and
                  Alexander Neitz and
                  Alexander Prokofiev and
                  Alexander Wei and
                  Allison Tam and
                  Ally Bennett and
                  Ananya Kumar and
                  Andre Saraiva and
                  Andrea Vallone and
                  Andrew Duberstein and
                  Andrew Kondrich and
                  Andrey Mishchenko and
                  Andy Applebaum and
                  Angela Jiang and
                  Ashvin Nair and
                  Barret Zoph and
                  Behrooz Ghorbani and
                  Ben Rossen and
                  Benjamin Sokolowsky and
                  Boaz Barak and
                  Bob McGrew and
                  Borys Minaiev and
                  Botao Hao and
                  Bowen Baker and
                  Brandon Houghton and
                  Brandon McKinzie and
                  Brydon Eastman and
                  Camillo Lugaresi and
                  Cary Bassin and
                  Cary Hudson and
                  Chak Ming Li and
                  Charles de Bourcy and
                  Chelsea Voss and
                  Chen Shen and
                  Chong Zhang and
                  Chris Koch and
                  Chris Orsinger and
                  Christopher Hesse and
                  Claudia Fischer and
                  Clive Chan and
                  Dan Roberts and
                  Daniel Kappler and
                  Daniel Levy and
                  Daniel Selsam and
                  David Dohan and
                  David Farhi and
                  David Mely and
                  David Robinson and
                  Dimitris Tsipras and
                  Doug Li and
                  Dragos Oprica and
                  Eben Freeman and
                  Eddie Zhang and
                  Edmund Wong and
                  Elizabeth Proehl and
                  Enoch Cheung and
                  Eric Mitchell and
                  Eric Wallace and
                  Erik Ritter and
                  Evan Mays and
                  Fan Wang and
                  Felipe Petroski Such and
                  Filippo Raso and
                  Florencia Leoni and
                  Foivos Tsimpourlas and
                  Francis Song and
                  Fred von Lohmann and
                  Freddie Sulit and
                  Geoff Salmon and
                  Giambattista Parascandolo and
                  Gildas Chabot and
                  Grace Zhao and
                  Greg Brockman and
                  Guillaume Leclerc and
                  Hadi Salman and
                  Haiming Bao and
                  Hao Sheng and
                  Hart Andrin and
                  Hessam Bagherinezhad and
                  Hongyu Ren and
                  Hunter Lightman and
                  Hyung Won Chung and
                  Ian Kivlichan and
                  Ian O'Connell and
                  Ian Osband and
                  Ignasi Clavera Gilaberte and
                  Ilge Akkaya},
  title        = {OpenAI o1 System Card},
  journal      = {CoRR},
  volume       = {abs/2412.16720},
  year         = {2024}
}

@misc{qwen2.5-1m,
    title = {Qwen2.5-1M: Deploy Your Own Qwen with Context Length up to 1M Tokens},
    url = {https://qwenlm.github.io/blog/qwen2.5-1m/},
    author = {Qwen Team},
    month = {January},
    year = {2025}
}

@article{qwen2.5,
      title={Qwen2.5-1M Technical Report}, 
      author={An Yang and Bowen Yu and Chengyuan Li and Dayiheng Liu and Fei Huang and Haoyan Huang and Jiandong Jiang and Jianhong Tu and Jianwei Zhang and Jingren Zhou and Junyang Lin and Kai Dang and Kexin Yang and Le Yu and Mei Li and Minmin Sun and Qin Zhu and Rui Men and Tao He and Weijia Xu and Wenbiao Yin and Wenyuan Yu and Xiafei Qiu and Xingzhang Ren and Xinlong Yang and Yong Li and Zhiying Xu and Zipeng Zhang},
      journal={arXiv preprint arXiv:2501.15383},
      year={2025}
}

@misc{still3,
    author={RUCAIBox STILL Team.},
    year   = {2025},
    title={STILL-3-1.5B-Preview: A 1.5B slow-thinking reasoning model continuously evolving through RL.},
    url = {https://github.com/RUCAIBox/Slow_Thinking_with_LLMs},
}

@misc{openthoughts,
  author = {Team, OpenThoughts},
  month = jan,
  title = {{Open Thoughts}},
  howpublished = {https://open-thoughts.ai},
  year = {2025}
}

@misc{deepscaler2025,
  title={DeepScaleR: Surpassing O1-Preview with a 1.5B Model by Scaling RL},
  author={Michael Luo and Sijun Tan and Justin Wong and Xiaoxiang Shi and William Y. Tang and Manan Roongta and Colin Cai and Jeffrey Luo and Li Erran Li and Raluca Ada Popa and Ion Stoica},
  year={2025},
  howpublished={\url{https://github.com/agentica-project/rllm}},
  note={Notion Blog},
  year={2025}
}

@misc{openai_o3_web,
    author={OpenAI},
    year   = {2024},
    title={OpenAI o3-mini},
    url = {https://openai.com/index/openai-o3-mini/},
}

@article{openai_o3,
  author       = {Ahmed El{-}Kishky and
                  Alexander Wei and
                  Andre Saraiva and
                  Borys Minaiev and
                  Daniel Selsam and
                  David Dohan and
                  Francis Song and
                  Hunter Lightman and
                  Ignasi Clavera Gilaberte and
                  Jakub Pachocki and
                  Jerry Tworek and
                  Lorenz Kuhn and
                  Lukasz Kaiser and
                  Mark Chen and
                  Max Schwarzer and
                  Mostafa Rohaninejad and
                  Nat McAleese and
                  o3 contributors and
                  Oleg M{\"{u}}rk and
                  Rhythm Garg and
                  Rui Shu and
                  Szymon Sidor and
                  Vineet Kosaraju and
                  Wenda Zhou},
  title        = {Competitive Programming with Large Reasoning Models},
  journal      = {CoRR},
  volume       = {abs/2502.06807},
  year         = {2025}
}

@article{deepseek_r1,
  author       = {DeepSeek{-}AI and
                  Daya Guo and
                  Dejian Yang and
                  Haowei Zhang and
                  Junxiao Song and
                  Ruoyu Zhang and
                  Runxin Xu and
                  Qihao Zhu and
                  Shirong Ma and
                  Peiyi Wang and
                  Xiao Bi and
                  Xiaokang Zhang and
                  Xingkai Yu and
                  Yu Wu and
                  Z. F. Wu and
                  Zhibin Gou and
                  Zhihong Shao and
                  Zhuoshu Li and
                  Ziyi Gao and
                  Aixin Liu and
                  Bing Xue and
                  Bingxuan Wang and
                  Bochao Wu and
                  Bei Feng and
                  Chengda Lu and
                  Chenggang Zhao and
                  Chengqi Deng and
                  Chenyu Zhang and
                  Chong Ruan and
                  Damai Dai and
                  Deli Chen and
                  Dongjie Ji and
                  Erhang Li and
                  Fangyun Lin and
                  Fucong Dai and
                  Fuli Luo and
                  Guangbo Hao and
                  Guanting Chen and
                  Guowei Li and
                  H. Zhang and
                  Han Bao and
                  Hanwei Xu and
                  Haocheng Wang and
                  Honghui Ding and
                  Huajian Xin and
                  Huazuo Gao and
                  Hui Qu and
                  Hui Li and
                  Jianzhong Guo and
                  Jiashi Li and
                  Jiawei Wang and
                  Jingchang Chen and
                  Jingyang Yuan and
                  Junjie Qiu and
                  Junlong Li and
                  J. L. Cai and
                  Jiaqi Ni and
                  Jian Liang and
                  Jin Chen and
                  Kai Dong and
                  Kai Hu and
                  Kaige Gao and
                  Kang Guan and
                  Kexin Huang and
                  Kuai Yu and
                  Lean Wang and
                  Lecong Zhang and
                  Liang Zhao and
                  Litong Wang and
                  Liyue Zhang and
                  Lei Xu and
                  Leyi Xia and
                  Mingchuan Zhang and
                  Minghua Zhang and
                  Minghui Tang and
                  Meng Li and
                  Miaojun Wang and
                  Mingming Li and
                  Ning Tian and
                  Panpan Huang and
                  Peng Zhang and
                  Qiancheng Wang and
                  Qinyu Chen and
                  Qiushi Du and
                  Ruiqi Ge and
                  Ruisong Zhang and
                  Ruizhe Pan and
                  Runji Wang and
                  R. J. Chen and
                  R. L. Jin and
                  Ruyi Chen and
                  Shanghao Lu and
                  Shangyan Zhou and
                  Shanhuang Chen and
                  Shengfeng Ye and
                  Shiyu Wang and
                  Shuiping Yu and
                  Shunfeng Zhou and
                  Shuting Pan and
                  S. S. Li},
  title        = {DeepSeek-R1: Incentivizing Reasoning Capability in LLMs via Reinforcement
                  Learning},
  journal      = {CoRR},
  volume       = {abs/2501.12948},
  year         = {2025}
}

@article{kimi_k1.5,
  author       = {Kimi Team and
                  Angang Du and
                  Bofei Gao and
                  Bowei Xing and
                  Changjiu Jiang and
                  Cheng Chen and
                  Cheng Li and
                  Chenjun Xiao and
                  Chenzhuang Du and
                  Chonghua Liao and
                  Chuning Tang and
                  Congcong Wang and
                  Dehao Zhang and
                  Enming Yuan and
                  Enzhe Lu and
                  Fengxiang Tang and
                  Flood Sung and
                  Guangda Wei and
                  Guokun Lai and
                  Haiqing Guo and
                  Han Zhu and
                  Hao Ding and
                  Hao Hu and
                  Hao Yang and
                  Hao Zhang and
                  Haotian Yao and
                  Haotian Zhao and
                  Haoyu Lu and
                  Haoze Li and
                  Haozhen Yu and
                  Hongcheng Gao and
                  Huabin Zheng and
                  Huan Yuan and
                  Jia Chen and
                  Jianhang Guo and
                  Jianlin Su and
                  Jianzhou Wang and
                  Jie Zhao and
                  Jin Zhang and
                  Jingyuan Liu and
                  Junjie Yan and
                  Junyan Wu and
                  Lidong Shi and
                  Ling Ye and
                  Longhui Yu and
                  Mengnan Dong and
                  Neo Zhang and
                  Ningchen Ma and
                  Qiwei Pan and
                  Qucheng Gong and
                  Shaowei Liu and
                  Shengling Ma and
                  Shupeng Wei and
                  Sihan Cao and
                  Siying Huang and
                  Tao Jiang and
                  Weihao Gao and
                  Weimin Xiong and
                  Weiran He and
                  Weixiao Huang and
                  Wenhao Wu and
                  Wenyang He and
                  Xianghui Wei and
                  Xianqing Jia and
                  Xingzhe Wu and
                  Xinran Xu and
                  Xinxing Zu and
                  Xinyu Zhou and
                  Xuehai Pan and
                  Y. Charles and
                  Yang Li and
                  Yangyang Hu and
                  Yangyang Liu and
                  Yanru Chen and
                  Yejie Wang and
                  Yibo Liu and
                  Yidao Qin and
                  Yifeng Liu and
                  Ying Yang and
                  Yiping Bao and
                  Yulun Du and
                  Yuxin Wu and
                  Yuzhi Wang and
                  Zaida Zhou and
                  Zhaoji Wang and
                  Zhaowei Li and
                  Zhen Zhu and
                  Zheng Zhang and
                  Zhexu Wang and
                  Zhilin Yang and
                  Zhiqi Huang and
                  Zihao Huang and
                  Ziyao Xu and
                  Zonghan Yang},
  title        = {Kimi k1.5: Scaling Reinforcement Learning with LLMs},
  journal      = {CoRR},
  volume       = {abs/2501.12599},
  year         = {2025}
}

@article{Logic-RL,
  author       = {Tian Xie and
                  Zitian Gao and
                  Qingnan Ren and
                  Haoming Luo and
                  Yuqian Hong and
                  Bryan Dai and
                  Joey Zhou and
                  Kai Qiu and
                  Zhirong Wu and
                  Chong Luo},
  title        = {Logic-RL: Unleashing {LLM} Reasoning with Rule-Based Reinforcement
                  Learning},
  journal      = {CoRR},
  volume       = {abs/2502.14768},
  year         = {2025},
  url          = {https://doi.org/10.48550/arXiv.2502.14768},
  doi          = {10.48550/ARXIV.2502.14768},
  eprinttype    = {arXiv},
  eprint       = {2502.14768},
  bibsource    = {dblp computer science bibliography, https://dblp.org}
}

@inproceedings{Self-playing-Adversarial,
  author       = {Pengyu Cheng and
                  Tianhao Hu and
                  Han Xu and
                  Zhisong Zhang and
                  Yong Dai and
                  Lei Han and
                  Nan Du and
                  Xiaolong Li},
  editor       = {Amir Globersons and
                  Lester Mackey and
                  Danielle Belgrave and
                  Angela Fan and
                  Ulrich Paquet and
                  Jakub M. Tomczak and
                  Cheng Zhang},
  title        = {Self-playing Adversarial Language Game Enhances {LLM} Reasoning},
  booktitle    = {Advances in Neural Information Processing Systems 38: Annual Conference
                  on Neural Information Processing Systems 2024, NeurIPS 2024, Vancouver,
                  BC, Canada, December 10 - 15, 2024},
  year         = {2024},
  url          = {http://papers.nips.cc/paper\_files/paper/2024/hash/e4be7e9867ef163563f4a5e90cec478f-Abstract-Conference.html},
  bibsource    = {dblp computer science bibliography, https://dblp.org}
}
